\documentclass[a4paper,fleqn]{cas-dc}

\usepackage[numbers,sort&compress]{natbib}
\usepackage{amsmath}
\usepackage{amssymb}
\usepackage{amsfonts}
\usepackage{mathtools}
\usepackage{graphicx}
\usepackage{url}
\usepackage{refcount}

\makeatletter
\g@addto@macro\UrlBreaks{\UrlOrds}
\makeatother
\usepackage{array}
\usepackage{booktabs}
\usepackage{multirow}
\usepackage{makecell}
\usepackage{siunitx}
\usepackage{subcaption}
\usepackage{algorithm}
\usepackage{algpseudocode}

\usepackage{amsthm}
\newtheorem{remark}{Remark}

\DeclareMathOperator*{\argmin}{arg\,min}
\usepackage{balance}

\graphicspath{{fig/}}

\begin{document}
\let\WriteBookmarks\relax
\def\floatpagepagefraction{1}
\def\textpagefraction{.001}

\shorttitle{Multimodal Trajectory Planning for Surface Vehicles using Turning Circle-based CBFs}
\shortauthors{C. Lee}

\title[mode = title]{Multimodal Trajectory Planning for Surface Vehicles using Turning Circle-based Control Barrier Functions}

\tnotemark[1]
\tnotetext[1]{This work was supported by the research grant of Kongju
National University in 2026.}

\author[1]{Changyu Lee}[type=editor,
                        auid=000,bioid=1,
                        orcid=0000-0003-1964-9101]
\ead{leeck@kongju.ac.kr}
\affiliation[1]{organization={Department of Mechanical and Automotive
                Engineering, Kongju National University},
                city={Cheonan},
                postcode={31080},
                country={Republic of Korea}}

\begin{abstract}
This paper presents a guide path-free multimodal trajectory planning framework for autonomous surface vehicles operating in dynamic environments. The proposed method integrates model predictive control (MPC) with a turning circle-based control barrier function (TC-CBF). Unlike conventional Euclidean distance-based CBFs (ED-CBFs), which evaluate safety solely based on proximity, the TC-CBF accounts for the nonholonomic motion and finite turning capability of a surface vehicle. Its geometric formulation identifies feasible avoidance regions according to the vehicle's turning circles and generates distinct left- and right-turning avoidance modes. These modes allow the optimization solver to explore and select topologically different trajectories without relying on globally planned guide paths, as required by many conventional multimodal planning approaches. By embedding the avoidance direction directly into the safety constraint, the proposed framework alleviates the local-minimum and deadlock problems of single-mode MPC while maintaining computational efficiency. Extensive simulations involving multiple moving vessels demonstrate that the proposed method achieves higher success rates, fewer safety violations, and smaller residual violations than single-mode baselines across all tested traffic densities.
\end{abstract}

% \begin{highlights}
% \item Guide path-free multimodal planner: side enumeration of the
% turning circle-based CBF spans distinct avoidance homotopy classes
% without any high-level sampling or graph-search planner.
% \item The candidate optimization problems share an identical structure and can be solved in parallel, enabling computationally efficient evaluation of multiple avoidance modes.
% \item The barrier function is evaluated using the turning circle of a ship governed by first-order steering dynamics, thereby incorporating finite turning capability directly into the safety constraint.
% \item Paired Monte Carlo simulations in structured maritime traffic demonstrate superior success and safety performance across varying traffic densities, with less severe safety violations than the baselines.
% \end{highlights}


\maketitle

\section{Introduction}\label{sec:intro}

Autonomous ships have attracted significant attention for
their potential to reduce human error, which remains the leading cause of
maritime accidents, and to improve the efficiency of transportation,
monitoring, and logistics operations.
A core requirement for their deployment is the capability to generate
safety-guaranteed and computationally efficient avoidance trajectories in
waters shared with multiple moving traffic ships.
A standard navigation architecture decomposes this problem into global route
planning over a static chart and local trajectory planning that reacts to
traffic.
Representative local planners include the dynamic window approach
\cite{fox1997dynamic}, artificial potential fields
\cite{zhang2023obstacle,zhai2024local}, velocity obstacles and their maritime
variants \cite{vo8_huang2018velocity,gvo_huang2019generalized}, and model
predictive control (MPC), which performs receding-horizon constrained
optimization subject to the vehicle model
\cite{wang2018design,mpc9_lee2023nonlinear,kim2023navigable,wu2025multivariate}.
In the maritime domain, MPC-based methods have been shown to handle traffic
rules and input constraints explicitly
\cite{mpc6_tsolakis2024model,scmpc4_tengesdal2022ship}, and control barrier
functions (CBFs) have been widely adopted to enforce collision-avoidance
constraints with formal forward-invariance properties
\cite{ames2016control,ames2019control,cbf3_thyri2020reactive,cbf4_xu2024safety,cbf6_basso2020safety,cbf8_wu2024constrained}.

In ship collision avoidance, trajectory generation involves both discrete
and continuous decisions: a planner must choose a passing side for each
relevant traffic ship while optimizing the corresponding state and control
trajectories.
Existing approaches handle these coupled decisions using velocity-obstacle
constructions
\cite{vo8_huang2018velocity,gvo_huang2019generalized,pvo_cho2020efficient},
metaheuristic searches \cite{xiao2024colregs}, scenario-based or
optimization-based MPC
\cite{scmpc4_tengesdal2022ship,mpc6_tsolakis2024model}, and CBF-based
reactive filters with ship-domain or turning-geometry modifications
\cite{cbf3_thyri2020reactive,thyri2022domain,lee2025efficient}.
Methods that prescribe a single passing side may fail when the resulting
maneuver becomes infeasible or when another vessel does not respond as
anticipated.
Conversely, methods that evaluate numerous sampled maneuvers or scenarios
can incur substantial computational cost.
This tradeoff motivates a planning framework that explicitly represents
distinct passing modes and evaluates them efficiently.

Gradient-based local planners also have a structural limitation. Once the
optimizer converges within one homotopy class, such as passing a traffic ship
on the port side, it cannot directly transition to a qualitatively different
avoidance route
\cite{bhattacharya2010search}.
In congested or symmetric encounters this manifests as local minima,
oscillatory maneuvers, or deadlock.
To address this, multimodal planning methods explicitly generate candidate
trajectories in distinct homotopy classes and select among them
\cite{de2024topology,li2025trust,adajania2022multi,chen2022interactive,bertipaglia2025multi,zhang2026homotopy}.
Most of these methods adopt a two-stage hierarchical pipeline in which a
high-level sampling or graph-search planner produces homotopy-distinct guide
paths that initialize parallel local optimizations.
While effective, the high-level stage adds computational overhead and design
complexity, and the initialization quality directly affects which modes are reachable.
Much of the existing multimodal planning literature has been developed for
mobile robots, autonomous vehicles, and aerial systems, whose maneuverability
allows geometric guide paths to be followed with relatively limited
deformation.
For ships, however, the large minimum turning radius and slow steering
response make turning capability an essential consideration already at the
trajectory-generation stage.
A geometrically collision-free path belonging to a desired homotopy class
may therefore be difficult or even infeasible for a ship to execute.
Accordingly, multimodal planning for ships should generate distinct avoidance
topologies while explicitly accounting for the vessel's finite turning
capability.

This paper proposes a guide path-free alternative built on the turning
circle-based CBF (TC-CBF) \cite{arxiv-tccbf,lee2025efficient}.
Instead of measuring safety by the Euclidean distance between the vehicle and
an obstacle, the TC-CBF measures the distance between the obstacle and the
center of the vehicle's instantaneous turning circle on a chosen side.
This construction has two consequences.
First, the turning limitation of the vehicle is embedded in the constraint,
so that the planned maneuvers remain executable by a ship whose minimum
turning radius spans several ship lengths.
Second, selecting the port or starboard turning circle assigns an avoidance
side to each obstacle. Enumerating the side choices for $M$ selected
obstacles yields $K=2^{M}$ structurally identical optimal control problems
(OCPs). These OCPs differ only in their side-parameter vectors, and their
solutions populate distinct homotopy classes without external guide paths.
Parallel computation has been employed in scenario-based maritime MPC to evaluate sampled control behaviors concurrently \cite{scmpc4_tengesdal2022ship}. In contrast, the proposed framework solves the complete trajectory optimization associated with each discrete avoidance topology in parallel.
The lowest-cost feasible mode is then executed in a receding-horizon manner.
In earlier work the TC-CBF served as a single-mode reactive safety
constraint \cite{arxiv-tccbf} and, combined with a rule-based side
selection, as a COLREGs-compliant reactive filter \cite{lee2025efficient};
both commit to one side per obstacle in advance.

The common structure of the $K$ problems enables efficient parallel
evaluation.
Because every mode shares one solver structure and differs only in a
parameter vector, all $K$ OCPs are dispatched as a single
OpenMP-parallelized batch \cite{dagum1998openmp}. Each solve performs a
real-time iteration \cite{diehl2005real} in acados
\cite{verschueren2022acados,frison2020hpipm}. The solver code is generated
once, and batch solving requires no per-mode compilation or code
duplication. Consequently, the wall-clock cost grows with
$\lceil K/P\rceil$ for $P$ threads rather than with $K$.
The proposed framework is evaluated through Monte Carlo simulations under
diverse multi-ship encounter scenarios to assess collision-avoidance safety,
goal-reaching performance, and computational feasibility.
The contributions of this study are summarized as follows:
\begin{itemize}
\item A guide path-free multimodal planning framework is proposed by embedding port- or starboard-side decisions directly into TC-CBF constraints. Enumerating the choices for the $M$ nearest ships yields $K=2^M$ OCPs representing distinct avoidance topologies.
\item The framework accounts for finite ship turning capability through TC-CBF and enables computationally efficient parallel evaluation using a common solver structure.
\item Monte Carlo simulations demonstrate that topology enumeration improves collision-avoidance safety and goal-reaching performance compared with heuristic side assignment while maintaining real-time feasibility.
\end{itemize}

The remainder of this paper is organized as follows.
Section~\ref{sec:pre} establishes the preliminaries.
Section~\ref{sec:method} details the guide path-free multimodal
planning framework.
Section~\ref{sec:results} presents the simulation results, and
Section~\ref{sec:conclusion} concludes the paper.

\section{Preliminaries}\label{sec:pre}

\subsection{Surface Vehicle Model}\label{sec:pre_model}
For trajectory planning purposes, a maneuvering model that captures the
dominant steering behavior at a low computational cost is preferred over a
full hydrodynamic model.
This study adopts the first-order Nomoto model
\cite{fossen2011handbook}, which is widely used for
autopilot and guidance design.
In the multimodal framework, every model evaluation is repeated across the
$K=2^{M}$ OCPs and the $N$ stages of each horizon. Consequently, the
marginal cost of a higher-fidelity model, such as a three-degree-of-freedom
MMG model with coupled sway-yaw hydrodynamics
\cite{yasukawa2015introduction}, is incurred $K N$ times per control period.
The Nomoto model captures the lagged and rate-limited heading response that
distinguishes a ship from a kinematic vehicle, using two scalar coefficients
that are routinely
identified from standard zig-zag or turning trials.
Unmodeled sway and speed-loss effects are absorbed by the safety margins
of the barrier formulation.
The state and input vectors are defined as
$\mathbf{x}=[x,\,y,\,\psi,\,u,\,r,\,\delta]^\top$ and
$\mathbf{u}=[a,\,\dot{\delta}]^\top$. Here, $(x,y)$ is the position in the
inertial frame, $\psi$ is the heading, $u$ is the forward speed, $r$ is the
yaw rate, and $\delta$ is the rudder angle.
The continuous-time dynamics are
\begin{equation}\label{eq:nomoto}
\dot{\mathbf{x}} \;=\;
f(\mathbf{x},\mathbf{u}) \;=\;
\begin{bmatrix}
u\cos\psi\\
u\sin\psi\\
r\\
a\\
\left(K_n\delta - r\right)/T_n\\
\dot{\delta}
\end{bmatrix},
\end{equation}
where $K_n$ and $T_n$ denote the Nomoto gain and time constant.
The rudder angle is included as a state so that both the rudder-angle limit
and the steering-gear rate limit are represented naturally:
\begin{equation}\label{eq:limits}
|\delta|\le\delta_{\max},\;\;
|\dot{\delta}|\le\dot{\delta}_{\max},\;\;
|a|\le a_{\max},\;\;
u_{\min}\le u\le u_{\max}.
\end{equation}
Two quantities derived from \eqref{eq:nomoto} are used throughout:
the saturated steady yaw rate and the corresponding minimum steady turning
radius,
\begin{equation}\label{eq:rmin}
r_{\text{ss}} = K_n\delta_{\max},\qquad
R_{\min}(u) = \frac{u}{r_{\text{ss}}}.
\end{equation}

\subsection{Control Barrier Functions}\label{sec:pre_cbf}
Consider a nonlinear system with state
$\mathbf{x}\in\mathbb{X}\subset\mathbb{R}^n$ and input
$\mathbf{u}\in\mathbb{U}\subset\mathbb{R}^m$.
Safety is described by a set
$\mathbb{S}=\{\mathbf{x}: h(\mathbf{x})\ge 0\}$ defined as the superlevel set
of a continuously differentiable function $h$, with boundary
$\partial\mathbb{S}=\{\mathbf{x}: h(\mathbf{x})=0\}$
\cite{ames2016control,ames2019control}.
The control objective is to render $\mathbb{S}$ forward invariant: any
trajectory starting in $\mathbb{S}$ remains in $\mathbb{S}$.
In the discrete-time setting used by predictive controllers, the CBF
condition is imposed as \cite{zeng2021safety}
\begin{equation}\label{eq:dcbf}
h(\mathbf{x}_{k+1}) \;\ge\; (1-\alpha)\,h(\mathbf{x}_k),
\qquad 0<\alpha\le 1,
\end{equation}
which permits the barrier value to decay at most geometrically toward the
boundary and thereby maintains safety even under a limited prediction
horizon.
For obstacle avoidance, the conventional choice is the Euclidean
distance-based CBF (ED-CBF)
\begin{equation}\label{eq:edcbf}
h_e(\mathbf{x}) = \big\|\,\mathbf{p}-\mathbf{o}\,\big\| - (o_r + R_s),
\end{equation}
where $\mathbf{p}=[x,y]^\top$ is the vehicle position, $\mathbf{o}$ and
$o_r$ are the obstacle center and radius, and $R_s$ is the vehicle safety
radius \cite{vulcano2022safe,jin2022collision,jian2023dynamic}.
The ED-CBF is undirected, penalizing proximity identically on every
side, and it ignores the turning limitation of the vehicle. This motivates
the turning circle-based construction adopted in
Section~\ref{sec:method_cbf}.

\subsection{Homotopy Classes and Multimodal Selection}\label{sec:pre_homotopy}
\begin{figure}
    \centering
    \includegraphics[width=0.92\linewidth]{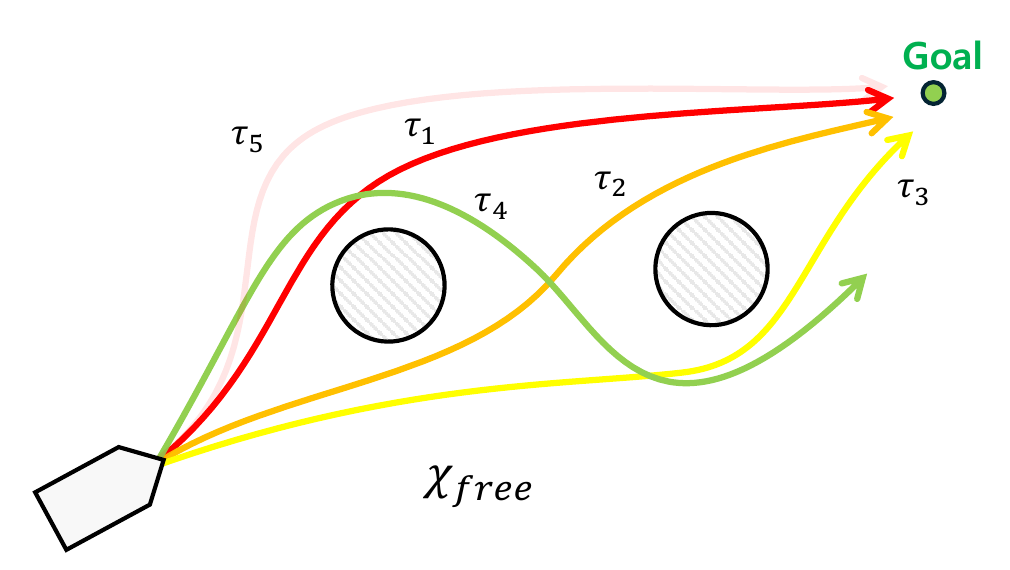}
    \caption{Homotopy classes around two obstacles: $\tau_1$-$\tau_4$
    realize the four side combinations and are topologically distinct,
    whereas $\tau_5$ can be deformed into $\tau_1$ and belongs to the same
    class.}
    \label{fig:homotopy}
\end{figure}

Two collision-free trajectories connecting the same endpoints belong to the
same homotopy class if one can be continuously deformed into the other
without intersecting obstacles \cite{bhattacharya2010search}, as
illustrated in Fig.~\ref{fig:homotopy}.
For $M$ obstacles whose passing side matters, the side choices generate
$K=2^{M}$ classes.

Gradient-based MPC converges within the class determined by its
initialization. Multimodal planning instead solves one OCP per class and
selects
\begin{equation}\label{eq:argmin_basic}
k^\ast=\argmin_{k\in\mathcal{K}_f} J_k^\ast,
\end{equation}
where $\mathcal{K}_f$ is the set of feasible modes and $J_k^\ast$ the
optimal cost of mode $k$ \cite{de2024topology,adajania2022multi}.
In the framework developed in Section~\ref{sec:method}, the classes are
produced by the side parameters alone, without high-level guide paths.

\section{Guide Path-Free Multimodal Planning}\label{sec:method}

\subsection{Problem Statement}\label{sec:method_problem}
The ego ship must track a waypoint route
$\{\mathbf{q}_m\}_{m=1}^{N_q}$ across open water populated by $N_t$ traffic
ships and reach the final waypoint without violating the safety zone of
any traffic ship.
The following working assumptions are made.
\emph{A1)} The positions, headings, and speeds of the traffic ships are
available at the control rate, e.g., through the automatic identification
system (AIS) or onboard tracking, and traffic motion is predicted with a
constant-velocity model over the horizon.
\emph{A2)} The operating area is free of extended boundaries such as
shorelines, shallow water, and channel limits. Confined waters would require
a boundary constraint in every mode and are outside the scope of this study.
Compact stationary
objects (moored vessels, buoys) may be present and are handled
identically to traffic ships with zero velocity under the prediction
model of A1.
\emph{A3)} The heading dynamics of the ego ship are adequately described
by the first-order Nomoto model \eqref{eq:nomoto} in the operating speed
range.
Under A1-A3, the task of the planner at each control period is to
determine both the topology of the avoidance maneuver (the passing
side for every relevant traffic ship) and the continuous trajectory
realizing it, within one sampling period.

\begin{figure}
    \centering
    \begin{subfigure}[t]{0.95\linewidth}
        \centering
        \includegraphics[width=\linewidth]{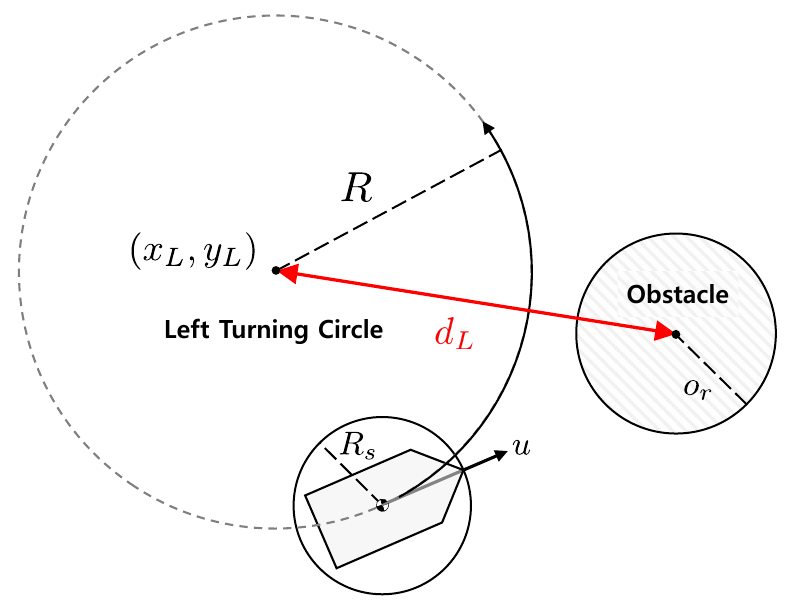}
        \caption{Port (left) turning circle.}
    \end{subfigure}\hfill
    \begin{subfigure}[t]{0.95\linewidth}
        \centering
        \includegraphics[width=\linewidth]{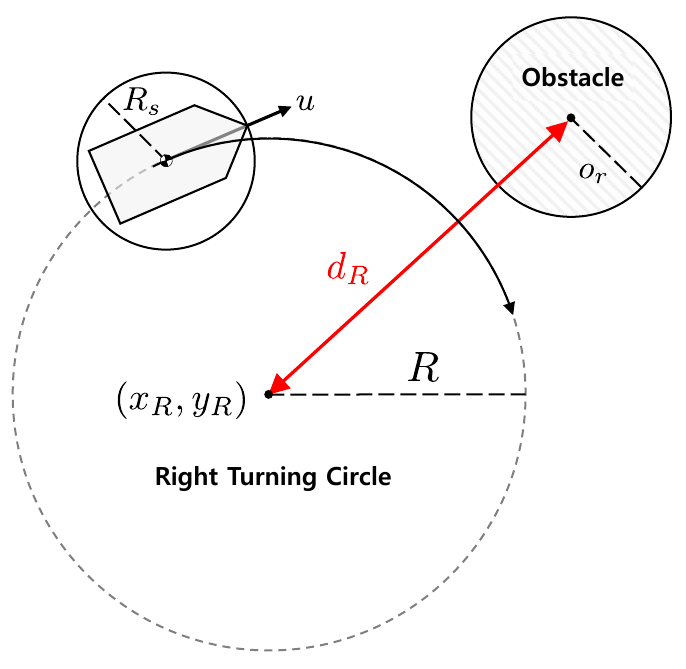}
        \caption{Starboard (right) turning circle.}
    \end{subfigure}
    \caption{Geometry of the TC-CBF \cite{arxiv-tccbf}: safety is measured between the obstacle and the center of the turning circle on the selected side, so that the corresponding escape turn remains available.}
    \label{fig:tccbf_geom}
\end{figure}

\begin{figure}
    \centering
    \begin{subfigure}[t]{0.99\linewidth}
        \centering
        \includegraphics[width=\linewidth]{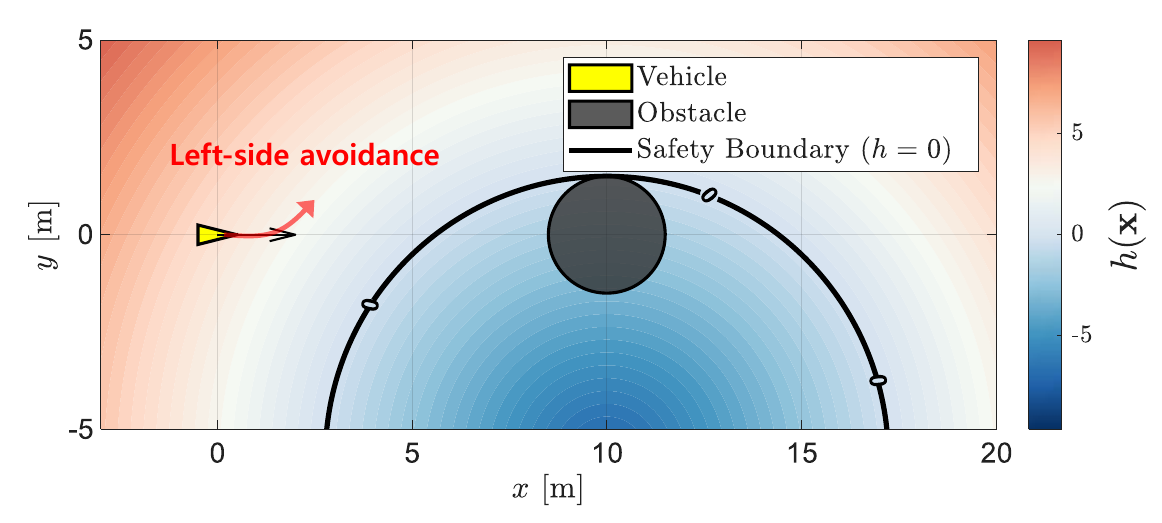}
        \caption{Value of the port-side barrier $h^{-1}$, which guides the vehicle to pass the obstacle on its port side.}
        \label{fig:ltc_field}
    \end{subfigure}
    \hfill
    \begin{subfigure}[t]{0.99\linewidth}
        \centering
        \includegraphics[width=\linewidth]{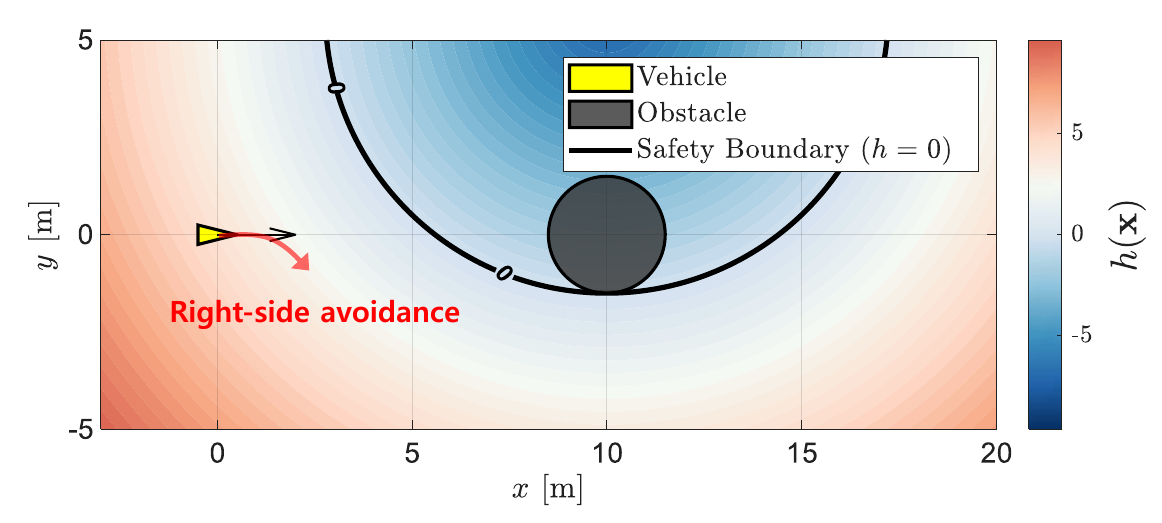}
        \caption{Value of the starboard-side barrier $h^{+1}$, which guides the vehicle to pass the obstacle on its starboard side.}
        \label{fig:rtc_field}
    \end{subfigure}
    \caption{Spatial distribution of the TC-CBF values $h^{\sigma}$ around an obstacle (black) at a fixed speed and heading. The region inside each zero level set is inadmissible under the corresponding side constraint, so the two barriers induce complementary passing directions from the same formulation.}
    \label{fig:tccbf_field}
\end{figure}

\subsection{TC-CBF for the ASV}\label{sec:method_cbf}
The safety constraint is built on the turning circle-based CBF (TC-CBF)
\cite{arxiv-tccbf,lee2025efficient}, which measures safety by the clearance
between an obstacle and the center of the vehicle's turning circle on a
chosen side, and which is instantiated here at the maneuvering limit of a
full-scale ship.
Let $\sigma\in\{+1,-1\}$ denote the avoidance side, where $\sigma=+1$
corresponds to the starboard (right) turning circle and $\sigma=-1$ to the
port (left) one.
For a turning radius $R$, the circle center is
\begin{equation}\label{eq:center}
\mathbf{p}_\sigma
= \mathbf{p} + R
\begin{bmatrix}
\cos(\psi-\sigma\pi/2)\\ \sin(\psi-\sigma\pi/2)
\end{bmatrix},
\end{equation}
and the barrier is
\begin{equation}\label{eq:tccbf}
h^{\sigma}(\mathbf{x}) = \big\|\,\mathbf{p}_\sigma-\mathbf{o}\,\big\|
- \big(o_r + R_s + R\big).
\end{equation}
The condition $h^{\sigma}\ge 0$ states that a hard turn toward side $\sigma$
sweeps a disk that remains clear of the obstacle; in other words, the
$\sigma$-side escape maneuver stays available.
Figure~\ref{fig:tccbf_geom} illustrates the construction for both sides.
Enforcing $h^{+1}$ therefore induces passing the obstacle with a starboard
turn, and $h^{-1}$ the opposite, so the side assignment directly selects the
homotopy class of the resulting trajectory.
Figure~\ref{fig:tccbf_field} shows the spatial distribution of the barrier
values around an obstacle at a fixed speed and heading. The zero level set
of each side barrier delimits the states from which the corresponding escape
turn is no longer collision-free, so the two inadmissible regions extend to
opposite sides of the obstacle.
Each region is anchored to the turning-circle center rather than the vehicle
position. Its shape consequently varies with the speed $u$ through $R$ and
with the relative orientation between vehicle and obstacle. The TC-CBF thus
provides a state-dependent, direction-selective constraint rather than a
static keep-out disk, and can represent both passing options.

The radius $R$ is the key quantity through which the vehicle's maneuvering limit enters the TC-CBF. In the kinematic setting of \cite{arxiv-tccbf}, $R=u/r_{\max}$, where the maximum yaw rate $r_{\max}$ is directly assignable. For a ship, however, the yaw rate is not a direct control input but is determined by the steering dynamics and actuator limits. Accordingly, \eqref{eq:center} and \eqref{eq:tccbf} are evaluated using the minimum sustainable turning radius $R_{\min}(u)=u/r_{\mathrm{ss}}$ from \eqref{eq:rmin}, corresponding to the steady-state hard-over turn achievable under the steering limit.
For a full-scale ship, this radius may span several ship lengths. By incorporating $R_{\min}$ into the barrier geometry, the TC-CBF accounts for the ship's finite turning capability and can reject passing-side assignments that are geometrically safe but dynamically infeasible. In contrast, an ED-CBF based solely on Euclidean separation cannot capture such directional maneuverability constraints.

\begin{remark}\label{rem:lag} 
The TC-CBF in \eqref{eq:tccbf} and the Nomoto dynamics in \eqref{eq:nomoto} capture complementary aspects of ship maneuverability. The TC-CBF is constructed using the steady-state minimum turning radius $R_{\min}(u)$ determined by the steering limit, thereby encoding the finite turning capability and directional feasibility of the ship into the safety constraint. Meanwhile, the Nomoto dynamics enter the OCP through the equality constraint \eqref{eq:ocp_dyn}, so the predicted trajectories explicitly account for the transient yaw response following a steering command. Over the prediction horizon, the optimizer can therefore anticipate the steering delay and initiate an avoidance maneuver accordingly. Thus, the TC-CBF determines feasible avoidance geometry based on the ship's turning capability, while the MPC realizes the maneuver subject to its transient steering dynamics. 
\end{remark}

Both the ego motion and the obstacle motion are propagated over one control interval when the discrete condition \eqref{eq:dcbf} is imposed. With sampling time $T_s$, the constraint attached to obstacle $j$ at stage
$k$ reads
\begin{equation}\label{eq:gcon}
\begin{split}
g_j^{\sigma_j}(\mathbf{x}_k)=\frac{1}{L}\Big[
&h^{\sigma_j}\!\big(\mathbf{x}_{k}^{+},\mathbf{o}_{j,k+1}\big)\\
&-(1-\alpha)\,
h^{\sigma_j}\!\big(\mathbf{x}_{k},\mathbf{o}_{j,k}\big)\Big]\ge 0,
\end{split}
\end{equation}
where $\mathbf{x}_k^{+}$ denotes the one-step prediction of the ego state and $\mathbf{o}_{j,k}=\mathbf{o}_j+\mathbf{v}_j\,T_s k$ is the constant-velocity prediction of obstacle $j$. Normalization by the ship length $L$ improves the numerical conditioning of the constraint.

\subsection{Waypoint Reference Generation}\label{sec:method_los}
The open-sea mission is specified as a sequence of waypoints $\{\mathbf{q}_m\}_{m=1}^{N_q}$, and the MPC reference is generated by lookahead-based LOS guidance \cite{lekkas2013line}. For the active route segment from $\mathbf{q}_m$ to $\mathbf{q}_{m+1}$ with path course $\chi_m$, the cross-track error of a point $\mathbf{p}$ is 
\begin{equation}\label{eq:cte}
e(\mathbf{p}) = -(x-q_{m,x})\sin\chi_m + (y-q_{m,y})\cos\chi_m,
\end{equation}
and the LOS course command is
\begin{equation}\label{eq:los}
\psi_{\text{los}}(\mathbf{p})
= \chi_m - \arctan\!\big(e(\mathbf{p})/\Delta\big),
\end{equation}
with lookahead distance $\Delta=5L$.
The reference trajectory $\{\mathbf{r}_i\}_{i=0}^{N}$ is constructed by
propagating a virtual point from the current position at the design speed
under \eqref{eq:los}, so that the reference itself bends back toward the
route and remains meaningful during avoidance detours.
Segment switching uses the circle-of-acceptance criterion augmented
with an along-track passing test: the next segment becomes active when
either the ship enters the acceptance circle of the upcoming waypoint or
its along-track coordinate exceeds the segment length.
This augmentation is required because an evasive maneuver can carry the ship
past a waypoint without entering its acceptance circle. With the circle
criterion alone, the guidance would continue tracking the previous segment.

\subsection{Multimodal Optimal Control Problem}\label{sec:method_ocp}
Let $\boldsymbol{\sigma}^{(k)}\in\{-1,+1\}^{N_o}$ collect the side
assignments of mode $k$ for the $N_o$ obstacles retained by the decision
module (Section~\ref{sec:method_decision}).
Each mode solves the same discrete-time OCP
\begin{subequations}\label{eq:ocp}
\begin{align}
\min_{\mathbf{x},\mathbf{u},\mathbf{s}}\;\;
&\sum_{i=0}^{N-1}\Big(\|\mathbf{x}_i-\mathbf{r}_i\|_Q^2
+\|\mathbf{u}_i\|_{R_d}^2+\rho_2\|\mathbf{s}_i\|_2^2\notag\\
&\qquad+\rho_1\|\mathbf{s}_i\|_1\Big)
+\|\mathbf{x}_N-\mathbf{r}_N\|_P^2\label{eq:ocp_cost}\\
\text{s.t.}\;\;
&\mathbf{x}_0=\mathbf{x}_{\text{init}},\\
&\mathbf{x}_{i+1}=f_d(\mathbf{x}_i,\mathbf{u}_i),\label{eq:ocp_dyn}\\
&\eqref{eq:limits}\ \text{on}\ \mathbf{x}_i,\ \mathbf{u}_i,\\
&g_j^{\sigma_j^{(k)}}(\mathbf{x}_i)\ge -s_{ij},\quad
s_{ij}\ge0,\label{eq:ocp_cbf}
\end{align}
\end{subequations}
for $i=0,\dots,N-1$ and $j=1,\dots,N_o$, where $f_d$ is the RK4 discretization of \eqref{eq:nomoto} and $\mathbf{r}_i$ is the LOS reference along the waypoint route \cite{lekkas2013line}. The CBF constraints are softened by slack variables $s_{ij}$, collected at each stage in $\mathbf{s}_i=[s_{i1},\dots,s_{iN_o}]^\top$. The mixed quadratic-linear term in \eqref{eq:ocp_cost}, with weights $\rho_2$ and $\rho_1$, penalizes these variables. 
Because $g_j^{\sigma_j}$ is normalized by
ship length in \eqref{eq:gcon}, the slacks are expressed in units of $L$.
The softening preserves feasibility in over-constrained instants, while the
constraint radius in \eqref{eq:tccbf} carries an additive buffer
$\Delta_b$ on $o_r$ (100\,m in Table~\ref{tab:params}) that slack
activation must consume before it can translate into a physical
safety-zone violation.
The weights $Q$, $R_d$, and $P$ penalize tracking error, input effort, and
terminal error, and $\|\mathbf{e}\|_W^2=\mathbf{e}^\top W\mathbf{e}$.
The speed-tracking weight is set high relative to the heading weight. With a
low speed weight, the optimizer can satisfy \eqref{eq:ocp_cbf} by slowing
behind a slower ship, which can cause overtaking deadlock. The higher weight
instead favors course alterations, which are more observable to surrounding
traffic than speed changes.

\begin{figure}
    \centering
    \begin{subfigure}[t]{0.99\linewidth}
        \centering
        \includegraphics[width=\linewidth]{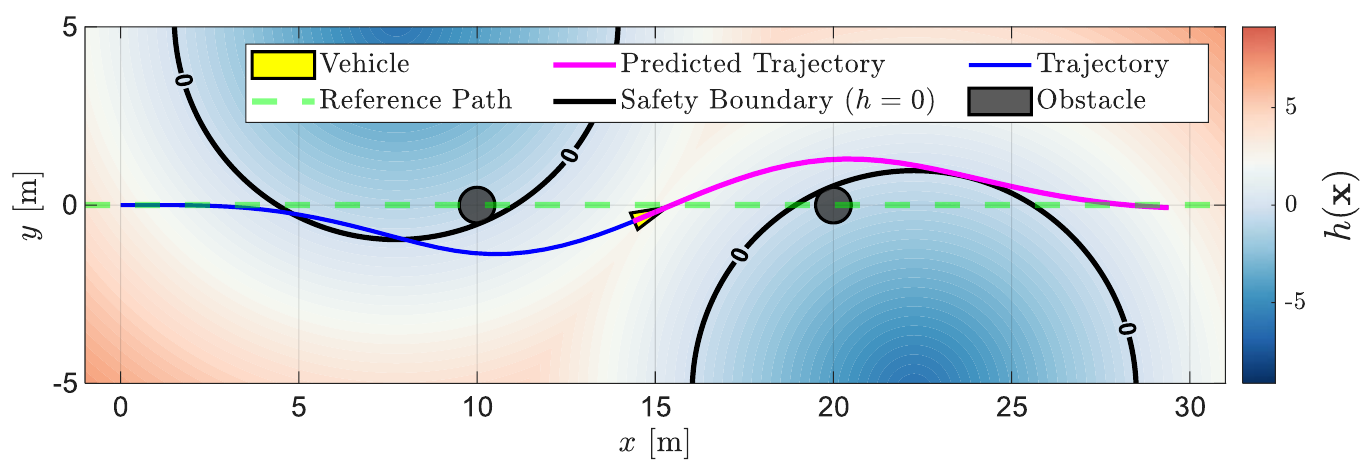}
        \caption{First homotopy trajectory ($\sigma_1={+}1$,
        $\sigma_2={-}1$).}
    \end{subfigure}
    \begin{subfigure}[t]{0.99\linewidth}
        \centering
        \includegraphics[width=\linewidth]{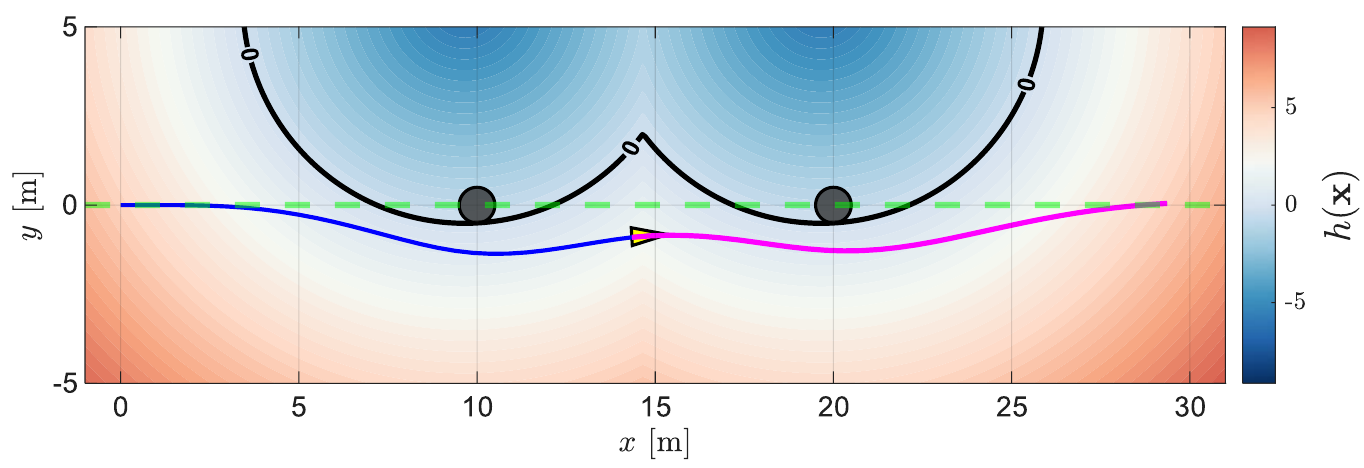}
        \caption{Second homotopy trajectory ($\sigma_1={+}1$,
        $\sigma_2={+}1$).}
    \end{subfigure}
    \begin{subfigure}[t]{0.99\linewidth}
        \centering
        \includegraphics[width=\linewidth]{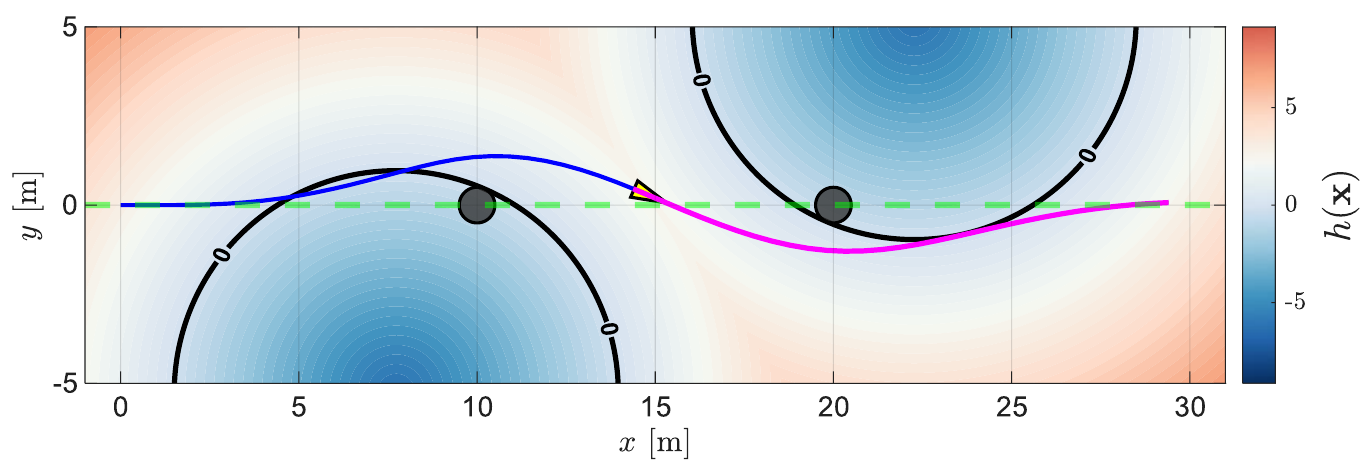}
        \caption{Third homotopy trajectory ($\sigma_1={-}1$,
        $\sigma_2={+}1$).}
    \end{subfigure}
    \begin{subfigure}[t]{0.99\linewidth}
        \centering
        \includegraphics[width=\linewidth]{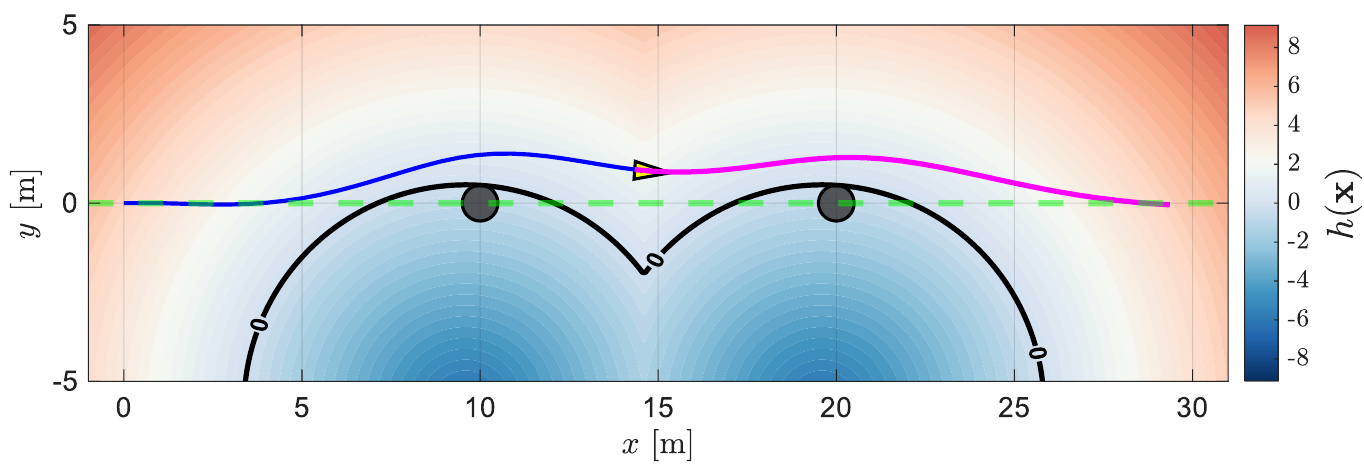}
        \caption{Fourth homotopy trajectory ($\sigma_1={-}1$,
        $\sigma_2={-}1$).}
    \end{subfigure}
    \caption{The four side combinations for two obstacles, each obtained
    by solving \eqref{eq:ocp} with a different side-parameter vector, span
    the four distinct homotopy classes without any guide path.}
    \label{fig:homotopy_traj}
\end{figure}

Figure~\ref{fig:homotopy_traj} illustrates the resulting mechanism for
two obstacles: enumerating the side pair
$(\sigma_1,\sigma_2)\in\{-1,+1\}^2$ and solving \eqref{eq:ocp} once for
each choice produces four trajectories that populate the four homotopy
classes of Fig.~\ref{fig:homotopy}.
No sampling, graph search, or guide path is involved, and the four problems
differ only in two constraint parameters, which is exactly the property
exploited by the batch solver of Section~\ref{sec:method_parallel}.

\subsection{Obstacle Selection and Mode Selection}
\label{sec:method_decision}
Enumerating side choices for every detected ship incurs exponential
computational cost. Conversely, constraining only the enumerated ships
would exclude nearby non-enumerated ships from the optimization.
The detected ships are ordered by surface distance, defined as the center
distance minus the obstacle radius, and divided into two proximity tiers.
The nearest $M$ ships form the branch set, for which all $2^{M}$ side
combinations are evaluated. The next-nearest $N_g$ ships form the
guard set and enter every mode with fixed, bearing-assigned sides. An
obstacle on the port bow is constrained by the starboard-side barrier and
vice versa. This preserves an escape maneuver away from the obstacle without
using additional branching capacity.
Hence every mode is aware of $N_o=M+N_g$ ships while only $K=2^{M}$
problems are solved.
Without the guard tier, a planner with $M=2$ would not account for the
third-nearest ship. This is problematic in the dense scenarios of
Section~\ref{sec:results}, where more than $M$ ships are frequently engaged
at once.
It also equalizes the comparison in Section~\ref{sec:results}: all
evaluated methods perceive the same $N_o$ ships and differ only in how the
avoidance sides of the branch set are determined.

The mode is selected once per control period after all $K$ OCPs are solved.
Let $\mathcal{K}_f$ denote the set of feasible modes and
$\hat{k}=\argmin_{k\in\mathcal{K}_f}J_k^\ast$ denote the lowest-cost
feasible mode. The selected mode is determined as
\begin{equation}\label{eq:select}
k^\ast=
\begin{cases}
\hat{k},
& \text{if } k_{\mathrm{prev}}\notin\mathcal{K}_f
\text{ or }
J_{\hat{k}}^\ast+\eta c_{\mathrm{med}}
<J_{k_{\mathrm{prev}}}^\ast,\\
k_{\mathrm{prev}},
& \text{otherwise},
\end{cases}
\end{equation}
where $c_{\mathrm{med}}$ is the median feasible cost and $\eta>0$
determines the switching hysteresis.
A mode is regarded as feasible when its real-time iteration returns either a
converged or iteration-limited solution. A mode whose QP fails numerically is
excluded from $\mathcal{K}_f$. Since the CBF constraints are softened, such
exclusion indicates solver failure rather than constraint incompatibility,
which instead appears as slack activation penalized in
\eqref{eq:ocp_cost}.
If all modes fail, the planner retains the previous mode and applies zero
input, thereby holding the rudder angle and speed. All solvers are then
re-initialized from the LOS reference at the next control period.
The hysteresis prevents unnecessary switching when multiple modes have
similar costs. Specifically, the planner switches to the lowest-cost mode
only when its cost improvement exceeds $\eta c_{\mathrm{med}}$, thereby
maintaining a stable passing intention.

\subsection{Parallel Batch Solution of the $K$ Modes}\label{sec:method_parallel}

The key structural property of the proposed formulation is that modes differ only through the parameter vector $\boldsymbol{\sigma}^{(k)}$ entering \eqref{eq:ocp_cbf}. The cost, dynamics, dimensions, and sparsity pattern of \eqref{eq:ocp} are identical for all $k$. Because the avoidance topology is encoded through constraint parameters rather than changes in the optimization structure, mode enumeration does not alter the solver structure.
Although the $K$ OCPs are mutually independent, efficient parallel execution requires a careful implementation. Naively generating and instantiating a separate solver for each mode would duplicate solver code and memory, while transferring data separately for each mode would introduce additional sequential overhead within each control step. To avoid this, the OCP solver code is generated only once, and $K$ solver instances are created using the same shared library. The $K$ OCPs are then dispatched as a single batch across $P$ OpenMP threads \cite{dagum1998openmp}. Each solver performs one real-time iteration \cite{diehl2005real} of sequential quadratic programming (SQP), with the resulting QP solved using a partially condensed interior-point method \cite{frison2020hpipm} in acados \cite{verschueren2022acados}.

The LOS reference, propagated obstacle parameters, and mode table are transferred through a single batched parameter interface, reducing the sequential data-preparation and transfer overhead before the parallel OCP solves. Denoting by $\bar{t}$ the mean single-OCP solve time, the wall-clock cost per control period is approximately 
\begin{equation} 
t_{\text{seq}}\approx K\,\bar{t} \qquad\text{versus}\qquad t_{\text{par}}\approx \big\lceil K/P\big\rceil\,\bar{t}+t_{\text{ovh}}, \end{equation} 
where $t_{\text{ovh}}$ denotes the scheduling and batch-processing overhead.

Three implementation aspects are worth noting. First, each mode is warm-started using its solution from the previous control period. If the QP solve fails for a particular mode, only that mode is re-initialized from the LOS reference at the next period.
Second, the prediction discretization and the replanning period are decoupled. The prediction horizon uses $T_s=20$ s, whereas the full batch is re-solved every $\Delta t_c=1$ s from the current state. At each replanning instant, one real-time iteration is performed for each mode \cite{diehl2005real}, and only the first control input is applied over the subsequent one-second interval. This coarse prediction grid keeps the 600 s horizon computationally tractable while allowing frequent replanning.
Third, the CBF constraints in \eqref{eq:gcon} are normalized by the ship length to improve numerical conditioning of the condensed QP. Algorithm~\ref{alg:mmmpc} summarizes the computations performed during one control period.

\begin{algorithm}[tbp]
\caption{Parallel multimodal MPC (one replanning period, 1\,Hz)}
\label{alg:mmmpc}
\begin{algorithmic}[1]
\State detect ships; order by surface distance
\State branch set $\gets$ nearest $M$; guard set $\gets$ next $N_g$
with bearing-assigned sides
\State build LOS reference $\{\mathbf{r}_i\}_{i=0}^{N}$ along the route
\State write batched parameters $\{\boldsymbol{\sigma}^{(k)}\}_{k=1}^{K}$,
obstacle predictions, reference
\State \textbf{parallel for} $k=1,\dots,K$: solve \eqref{eq:ocp}
\Comment{OpenMP}
\State select $k^\ast$ by \eqref{eq:select}; apply
$\mathbf{u}_0^{(k^\ast)}$ for $\Delta t_c$
\end{algorithmic}
\end{algorithm}

\section{Validation Results}\label{sec:results}

\subsection{Simulation Setup}\label{sec:results_setup}
The ego ship is a VLCC-class vessel with length $L=320$\,m and design speed $U=7.97$\,m/s, comparable in scale to the KVLCC2 benchmark hull \cite{yasukawa2015introduction}. Representative Nomoto coefficients were chosen to reproduce VLCC-class turning performance: the steady turning radius at the maximum rudder angle
$\delta_{\max}=35^\circ$ was set to $2.0L$, giving $K_n=0.0204$\,s$^{-1}$,
and the nondimensional time constant $T_n'=2.5$ gives $T_n=100.4$\,s.
Table~\ref{tab:params} lists the remaining parameters.
The traffic ships follow constant-velocity models; their positions and
velocities are assumed available (e.g., from AIS), consistent with the
constant-velocity prediction embedded in \eqref{eq:gcon}.
A safety-zone violation is declared when the center distance to any
traffic ship falls below the combined safety radius ($R_s+o_r$). The CBF
radius includes a 100\,m buffer, allowing the soft constraint to use this
margin before a safety-zone violation occurs.
All computations run under Ubuntu~24.04 (WSL2) on a desktop with an Intel
Core Ultra~7 270K Plus CPU and 32\,GB of RAM. The CPU has 24 physical cores
(eight performance and sixteen efficiency cores), no simultaneous
multithreading, and a maximum clock speed of 3.7\,GHz. The acados/HPIPM solver
stack is compiled with OpenMP.
The prediction horizon is $N T_s=600$\,s, corresponding to approximately
4.8\,km at the design speed. The closed-loop
planner updates every $\Delta t_c=1$\,s, as described in
Section~\ref{sec:method_parallel}.

\begin{table}[tbp]
\centering
\caption{Simulation and controller parameters.}
\label{tab:params}
\setlength{\tabcolsep}{5.2pt}
\renewcommand{\arraystretch}{1.25}
\begin{tabular}{llll}
\toprule
\textbf{Symbol} & \textbf{Value} & \textbf{Symbol} & \textbf{Value}\\
\midrule
$L$ & 320\,m & $U$ & 7.97\,m/s\\
$K_n$ & 0.0204\,s$^{-1}$ & $T_n$ & 100.4\,s\\
$\delta_{\max}$ & $35^\circ$ & $\dot{\delta}_{\max}$ & $3^\circ$/s\\
$a_{\max}$ & 0.02\,m/s$^2$ & $u_{\min},u_{\max}$ & 3.0, 9.0\,m/s\\
$R_s$ & 500\,m & CBF buffer & 100\,m\\
$T_s$ & 20\,s & $N$ & 30\\
$\Delta t_c$ & 1\,s & sensing range & 8\,km\\
$\alpha$ & 0.3 & $\eta$ & 0.25  \\
$M$ (branch) & 4 & $N_g$ (guard) & 2\\
 $\rho_1$ & $10^{4}$ & $\rho_2$ & $10^{6}$ \\
$Q$ & \multicolumn{3}{l}{$\mathrm{diag}(10^{-4},10^{-4},500,50,0,0)$}\\
$P$ & \multicolumn{3}{l}{$\mathrm{diag}(10^{-4},10^{-4},500,50,0,0)$}\\
$R_d$ & \multicolumn{3}{l}{$\mathrm{diag}(10^{3},5\times10^{4})$}\\
\bottomrule
\end{tabular}
\end{table}

The mission is a fixed 33.7\,km waypoint route with two course changes.
It includes slower vessels along the route (overtaking encounters), reciprocal vessels in a laterally offset opposing lane (head-on encounters), and vessels crossing from port or starboard. 
Their arrival times create encounters with the ego ship. Stationary obstacles, representing moored vessels or buoys, are placed within 0.5-1.3\,km of the route. Dynamic ships sail at 3.0-6.5\,m/s with safety radii of
400-600\,m; static obstacles have radii of 250-450\,m.
Three density levels are used: \textbf{D1} (8 dynamic $+$ 3 static), \textbf{D2} (12 dynamic
$+$ 4 static), and \textbf{D3} (16 dynamic $+$ 5 static). Encounter times are
distributed over the transit to maintain sustained congestion.
Traffic ships maintain constant course and speed. Vessels in the same lane
share a common speed, and the generator prevents their safety zones from
overlapping during the simulation. Thus, only the ego ship must actively
resolve conflicts.
A trial is a success if the ship completes the route without any
safety-zone violation within twice the nominal transit time.

All compared methods perceive the same $N_o=6$ nearest ships and differ only in how
avoidance sides are determined, isolating the value of topology branching:
\begin{itemize}
\item \textbf{ED-CBF} ($K{=}1$): six undirected distance-based constraints \eqref{eq:edcbf}, representing conventional CBF-MPC \cite{zeng2021safety,jian2023dynamic}.
\item \textbf{TC-CBF single} ($K{=}1$): six TC-CBF constraints \eqref{eq:tccbf} with bearing-assigned sides and no branching \cite{lee2025efficient, arxiv-tccbf}.
\item \textbf{Proposed ($K{=}16$)}: branch set 4, guard set 2.
\end{itemize}

\begin{figure}
    \centering
    \begin{subfigure}[t]{0.99\linewidth}
        \centering
        \includegraphics[width=\linewidth]{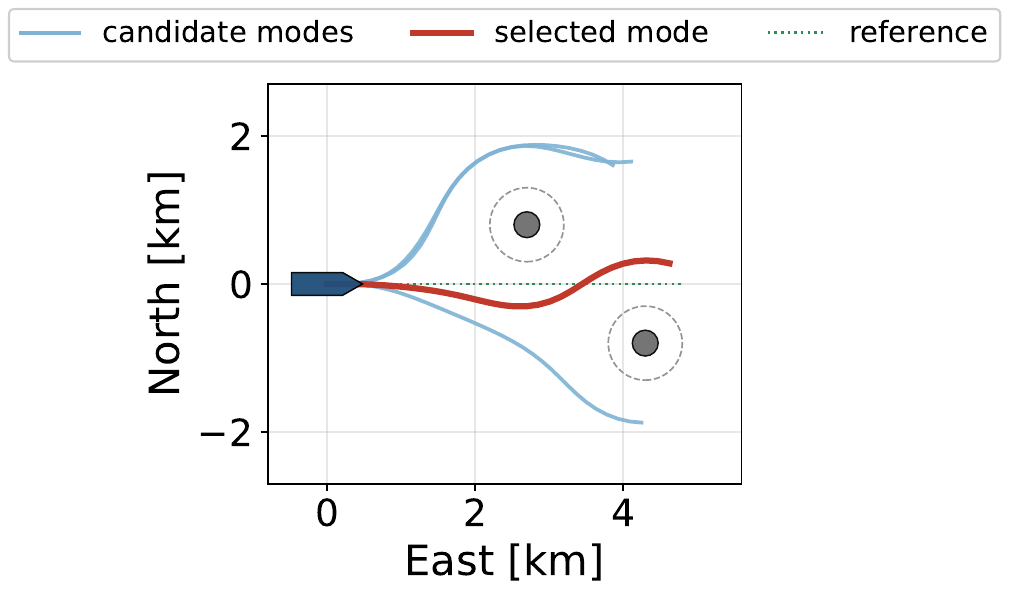}
        \caption{Two static obstacles (gray disks with their combined
        keep-out radii dashed).}
        \label{fig:fan_static}
    \end{subfigure}
    \hfill
    \begin{subfigure}[t]{0.99\linewidth}
        \centering
        \includegraphics[width=\linewidth]{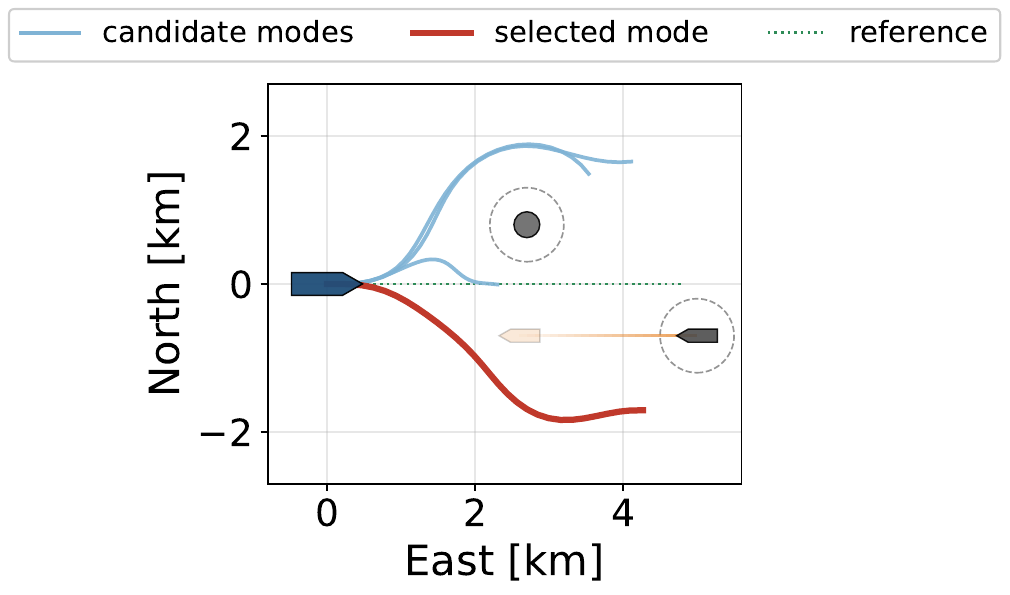}
        \caption{One static obstacle and one ship closing head-on, drawn
        as a ship glyph with its constant-velocity prediction over the
        600\,s horizon.}
        \label{fig:fan_dyn}
    \end{subfigure}
    \caption{Topology enumeration with two branched obstacles: the $K=2^2=4$ side combinations generate the candidate trajectories (light blue), with the selected mode shown in red. Candidates for which the second side constraint is inactive coincide.}
    \label{fig:fan}
\end{figure}

\subsection{Topology Enumeration and Candidate Modes}
\label{sec:results_fanout}
Figure~\ref{fig:fan} illustrates the mechanics of the side enumeration in
two open-loop snapshots, each with two branched obstacles and hence
$K=2^2=4$ candidate modes.
In the static arrangement (Fig.~\ref{fig:fan_static}), the four side
combinations span approximately $\pm2$\,km laterally. The selected mode
passes between the two obstacles with the smallest route deviation.
When the second obstacle is a ship closing head-on
(Fig.~\ref{fig:fan_dyn}), the enumeration uses the constant-velocity
prediction rather than the current position. The candidates separate
according to which side of the predicted track the ego ship will
pass. Because the route between the two obstacles is no longer available,
the selector commits early to a single-sided maneuver.
Neither case involves a guide path, sampling, or graph search, and only the
side-parameter vector changes between the four problems.

\subsection{Density Sweep in Structured Maritime Traffic}
\label{sec:results_density}
This experiment examines whether increasing the branching capacity improves
navigation performance as traffic density increases and quantifies the
associated computational cost.
One hundred paired Monte Carlo trials were run at each density level; all
methods face identical traffic per trial.
Closed-loop videos of every realization shown in this section are
available on the supplementary page.\footnote{\label{fn:videos}\url{https://leechangyu95.github.io/Multimodal-Trajectory-Planning-for-Surface-Vehicles/}}
Because a safety-zone infringement indicates entry into a virtual keep-out region rather than an actual hull collision, each trial is simulated to the end of the route even after the first infringement occurs.
The success statistics are unaffected because a trial is classified as a
violation at the first zone breach. 

\begin{table*}[tbp]
\centering
\caption{Paired Monte Carlo results across traffic densities (100 trials per cell), with all methods perceiving the same six nearest ships. Each cell reports the success and safety-violation rates (\%), followed by the median safety-zone penetration depth among violating trials; the remaining trials are timeouts, which neither violate a safety zone nor finish the route within twice the nominal transit time. Penetration depth is measured by continuing each simulation after the first violation.}
\label{tab:density}
\setlength{\tabcolsep}{3.4pt}
\renewcommand{\arraystretch}{1.3}
\begin{tabular}{lcccccc}
\toprule
& \multicolumn{2}{c}{\textbf{D1} (8 dyn)}
& \multicolumn{2}{c}{\textbf{D2} (12 dyn)}
& \multicolumn{2}{c}{\textbf{D3} (16 dyn)}\\
\cmidrule(lr){2-3}\cmidrule(lr){4-5}\cmidrule(lr){6-7}
\textbf{Method} & suc./viol. & $\bar{d}$ [m] & suc./viol. & $\bar{d}$ [m]
& suc./viol. & $\bar{d}$ [m]\\
\midrule
ED-CBF ($K{=}1$) & 40\,/\,60 & 250 & 31\,/\,69 & 229 & 20\,/\,80 & 266\\
TC-CBF single ($K{=}1$) & 76\,/\,24 & 608 & 77\,/\,23 & 646 & 65\,/\,31 &
598\\
Proposed $K{=}16$ & \textbf{99\,/\,1} & \textbf{219} & \textbf{97\,/\,3} &
\textbf{127} & \textbf{86\,/\,10} & \textbf{19}\\
\bottomrule
\end{tabular}
\end{table*}

\begin{figure}
    \centering
    \includegraphics[width=\linewidth]{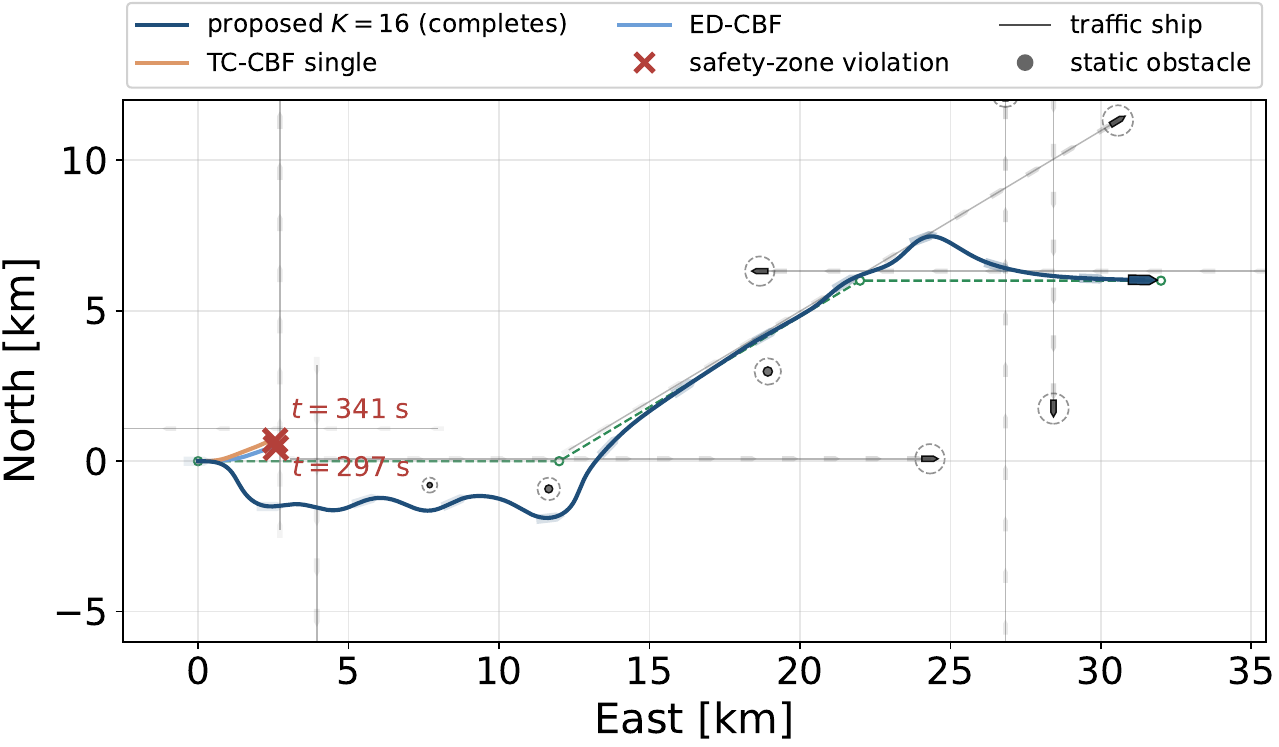}
    \caption{Paired realization at \textbf{D1} (8 dynamic $+$ 3 static; all three
    planners face identical traffic). The undirected ED-CBF is stopped by
    the opening encounter group ($t=297$\,s) and the single TC-CBF
    shortly after ($t=341$\,s); the proposed $K{=}16$ planner re-commits
    its topology 8 times and completes the route
    (videos\protect\footnotemark[\getrefnumber{fn:videos}]).}
    \label{fig:show8}
\end{figure}

\begin{figure}
    \centering
    \includegraphics[width=\linewidth]{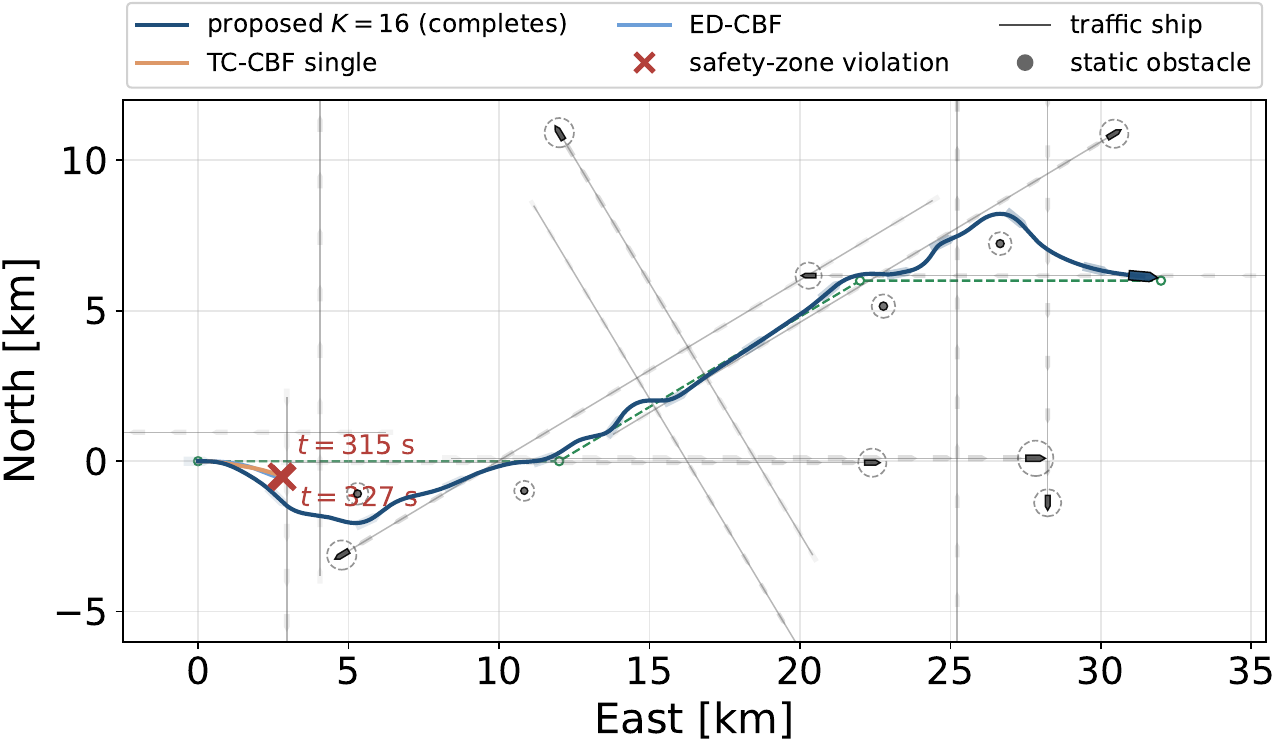}
    \caption{Paired realization at \textbf{D2} (12 dynamic $+$ 4 static): the
    single TC-CBF fails at $t=315$\,s and the ED-CBF at $t=327$\,s, while
    the proposed planner re-commits 17 times and completes the route
    (videos\protect\footnotemark[\getrefnumber{fn:videos}]).}
    \label{fig:show12}
\end{figure}

\begin{figure}
\centering
    \includegraphics[width=\linewidth]{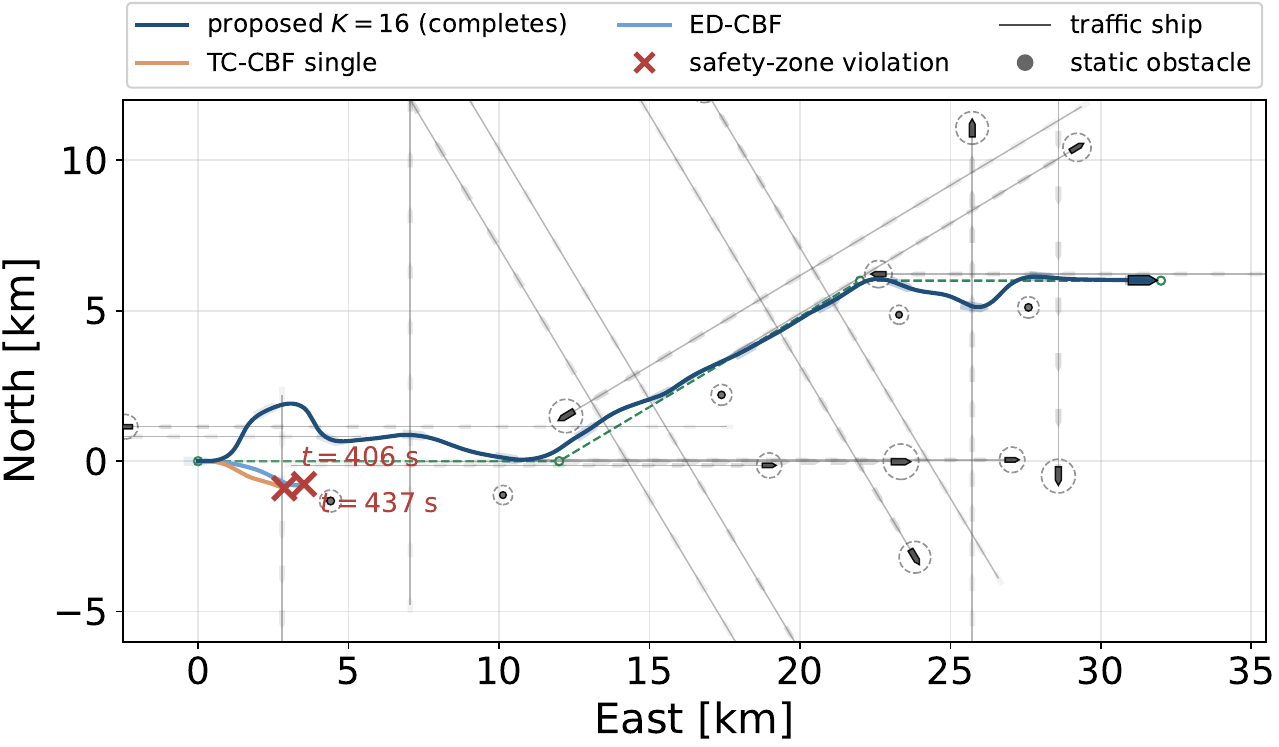}
    \caption{Paired realization at \textbf{D3} (16 dynamic $+$ 5 static): both
    single-mode baselines fail in the opening group ($t=406$\,s and
    $t=437$\,s), while the proposed planner re-commits its topology 26
    times under sustained congestion and completes the route
    (videos\protect\footnotemark[\getrefnumber{fn:videos}]).}
    \label{fig:show16}
\end{figure}

\begin{figure}
    \centering
    \includegraphics[width=0.9\linewidth]{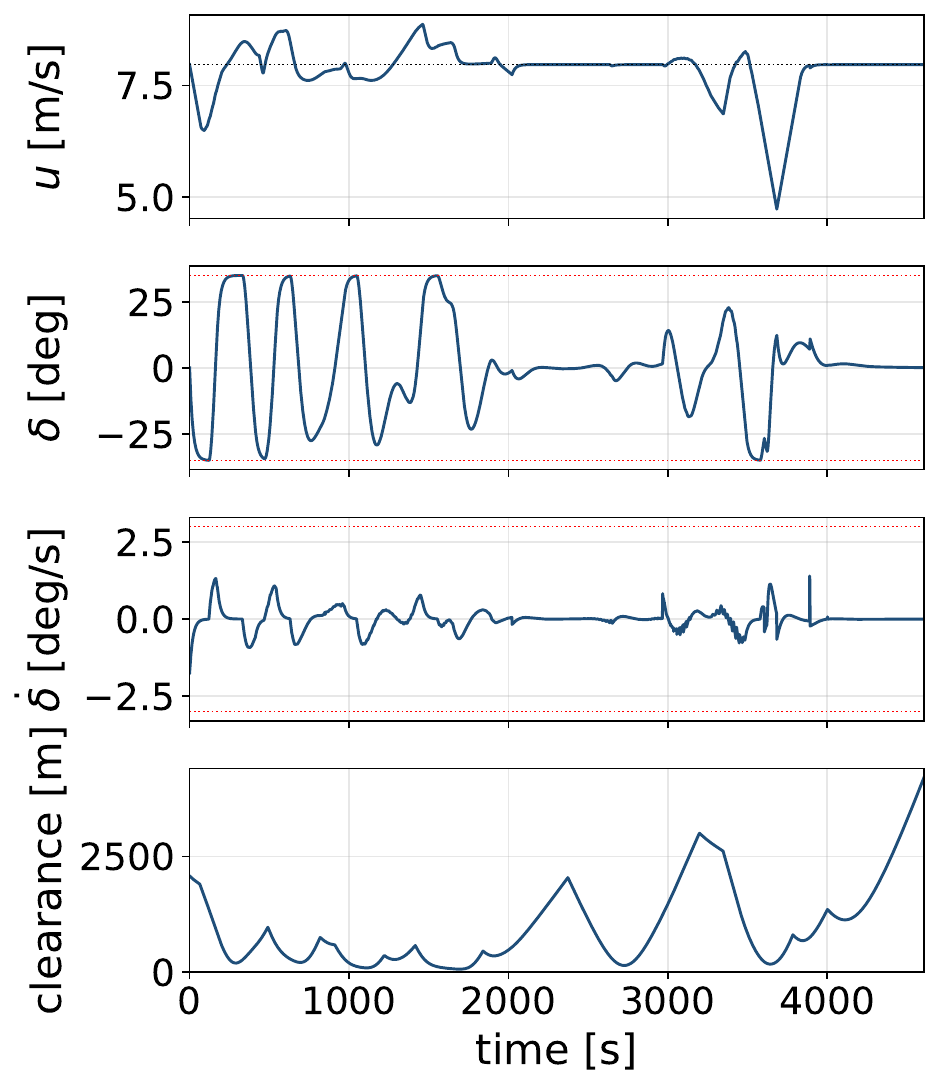}
    \caption{Time histories of the proposed method for the \textbf{D1}
    realization of Fig.~\ref{fig:show8}: speed, rudder angle,
    rudder-rate input, and clearance beyond the combined safety radius.}
    \label{fig:ts8}
\end{figure}

\begin{figure}
    \centering
    \includegraphics[width=0.9\linewidth]{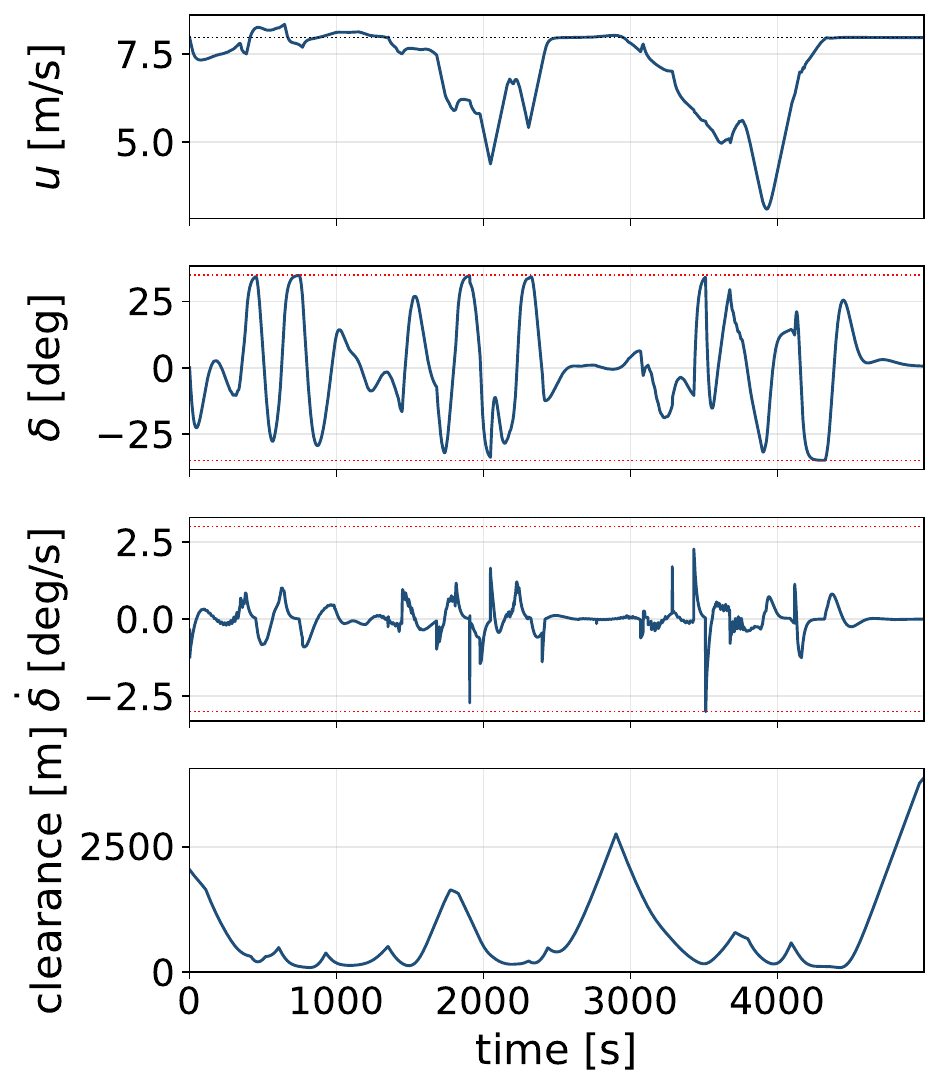}
    \caption{Time histories of the proposed method for the \textbf{D2}
    realization of Fig.~\ref{fig:show12}.}
    \label{fig:ts12}
\end{figure}

\begin{figure}
    \centering
    \includegraphics[width=0.9\linewidth]{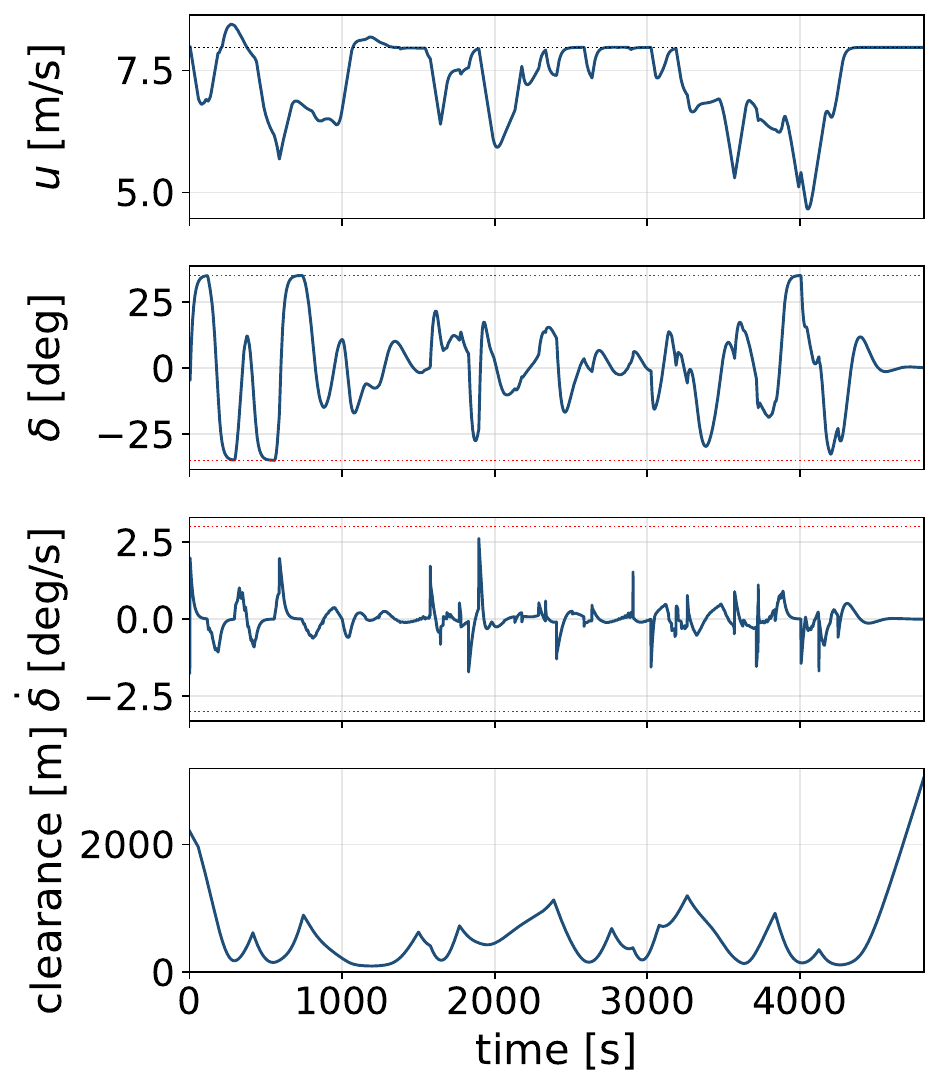}
    \caption{Time histories of the proposed method for the \textbf{D3}
    realization of Fig.~\ref{fig:show16}.}
    \label{fig:ts16}
\end{figure}

\begin{figure}
    \centering
    \includegraphics[width=\linewidth]{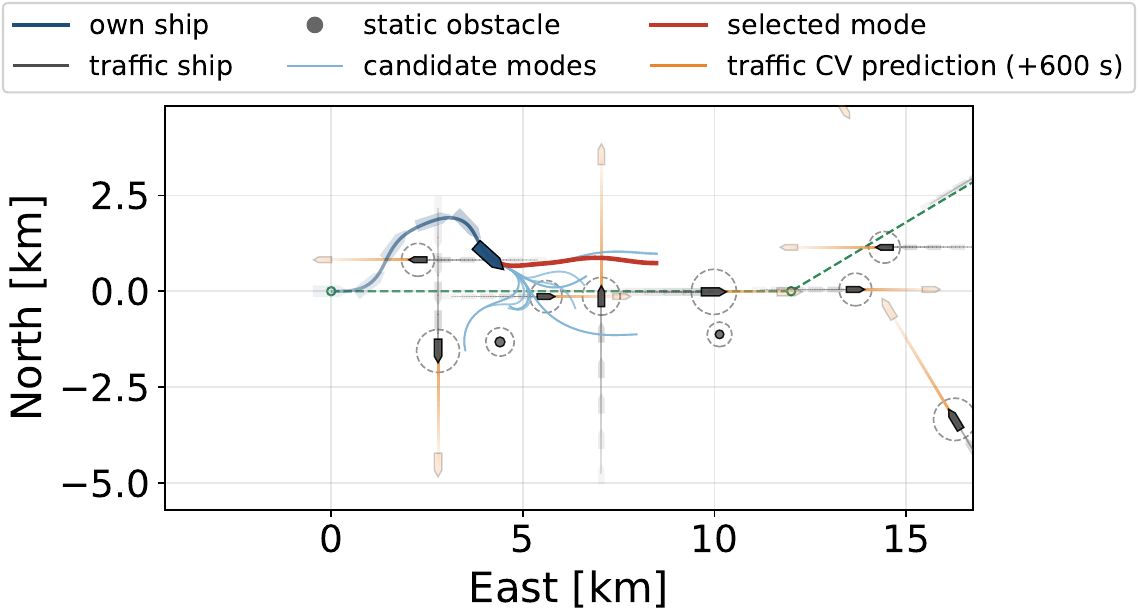}
    \caption{Mode divergence during the \textbf{D3} transit of
    Fig.~\ref{fig:show16}, at $t=752$\,s and centered on the encounter
    cluster: six obstacles are decision-relevant, the nearest 1.4\,km
    ahead. All sixteen candidates are feasible and their terminal states
    spread 5.2\,km laterally (5.6\,km across), fourteen passing south of
    the ego ship and two north. The selected mode overtakes the slow ship
    ahead on its northern side with 0.09\,km of clearance, whereas the
    bearing-assigned combination that the single TC-CBF baseline follows
    passes the same ship to the south and, while also collision-free
    here, carries 221 times the mode cost, so the commitment is
    set by the traffic pattern as a whole rather than by the nearest ship
    alone.}
    \label{fig:d3snap}
\end{figure}

Figures~\ref{fig:show8}-\ref{fig:show16} show one paired realization at
each density: the three planners face identical traffic, and the
corresponding time histories of the proposed method are given in
Figs.~\ref{fig:ts8}-\ref{fig:ts16}.
The performance difference increases with traffic density.
Both single-mode baselines fail during the opening encounter group. The
ED-CBF and single TC-CBF fail at 297\,s and 341\,s, respectively, with eight
dynamic ships, and at 437\,s and 406\,s with sixteen dynamic ships. Each
single-mode planner lacks an alternative when its initial topology becomes
infeasible.
The proposed planner clears the same encounter group and completes the route
in all three cases. It changes its selected topology 8, 17, and 26 times as the branch
set changes, with minimum clearances of 60, 82, and 92\,m beyond the combined
safety radius, respectively.
The time histories confirm that the resulting maneuvers satisfy the input
limits. The rudder saturates only during evasive maneuvers, and the
rudder-rate input remains within its bound. The speed also remains near the
design value.
Figure~\ref{fig:d3snap} illustrates the multimodal mechanism for a cluster
of four branch ships and two guarded obstacles. The candidate set
simultaneously maintains northern,
southern, and gap-threading alternatives, and the commitment is made by
cost over the whole traffic pattern rather than against the nearest ship
only.
At that instant, the bearing-assigned combination remains collision-free but
has 221 times the cost of the selected mode. The heuristic directs the ego
ship to the side of the overtaken vessel already occupied by a static
obstacle and a westbound ship. The enumerated selector therefore chooses a
different combination before any individual encounter becomes critical.

\begin{figure}
    \centering
    \includegraphics[width=\linewidth]{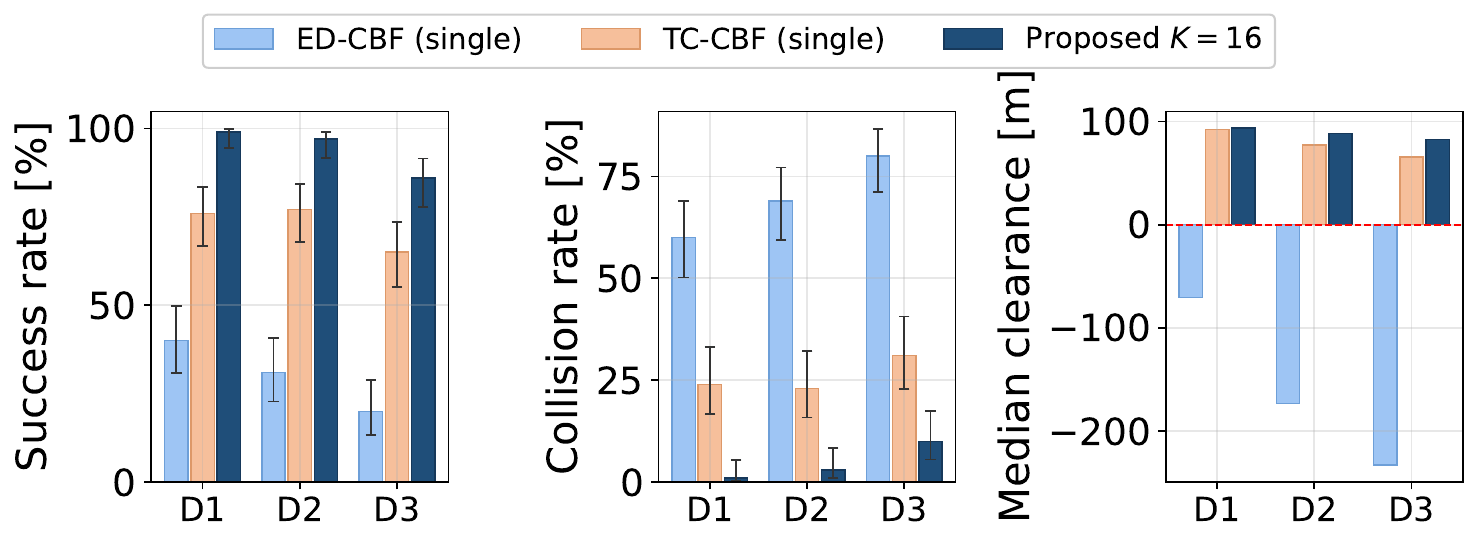}
    \caption{Monte Carlo outcome versus density: success rate (left),
    safety-violation rate (center), and median minimum clearance (right)
    for the three methods of Table~\ref{tab:density}.
    Error bars on the two rate panels are 95\% confidence intervals ($n{=}100$).}
    \label{fig:mcbars}
\end{figure}

Table~\ref{tab:density} summarizes the paired comparison, while
Fig.~\ref{fig:mcbars} visualizes the same results together with the median
clearances.
Three observations can be made.
First, the proposed $K{=}16$ configuration achieves the highest success rate and the lowest violation rate across all traffic densities. Its success rates are 99\%,
97\% and 86\%, compared with 20-77\% for the single-mode baselines. 
Second, the performance of the undirected ED-CBF degrades as traffic density
increases, with the success rate decreasing from 40\% to
20\%. 
In contrast, encoding the passing direction through the TC-CBF improves the success rate to 65-77\% under the same perception budget, indicating the benefit of direction-aware safety constraints in multi-ship encounters.
Third, a single directional topology remains insufficient in more complex traffic. The bearing-assigned TC-CBF achieves success rates of 76\%, 77\%, and 65\%, whereas enumerating multiple topologies increases them to 99\%, 97\%, and 86\%, respectively. This consistent improvement shows that maintaining multiple avoidance alternatives is particularly beneficial when several ships are simultaneously decision-relevant.

\begin{figure}
    \centering
    \includegraphics[width=\linewidth]{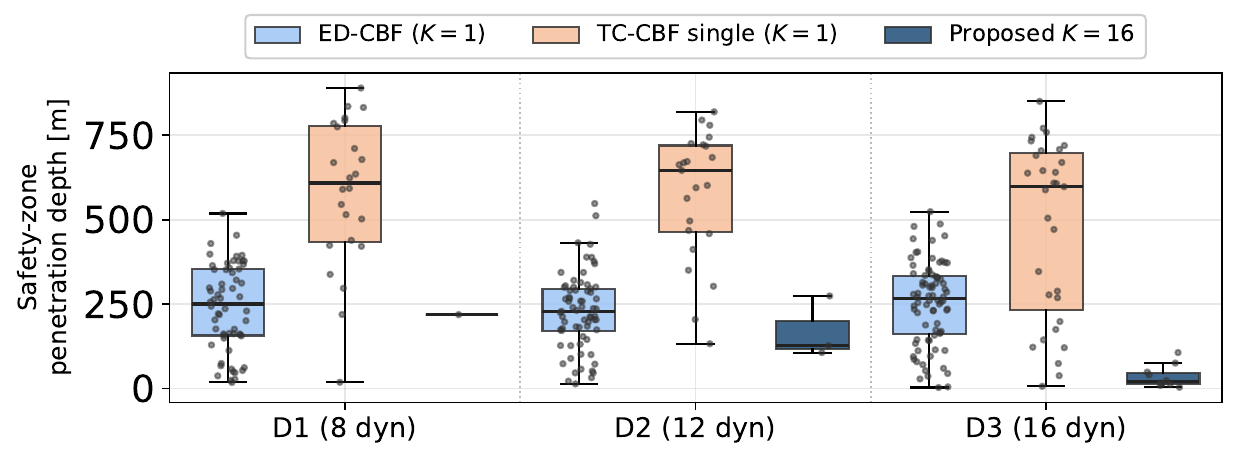}
    \caption{Distribution of the safety-zone penetration depth over the
    violating trials of Table~\ref{tab:density} (boxes: median and
    interquartile range; dots: individual trials). Counts differ per box
    because the methods violate at different rates; the proposed planner
    contributes 1, 3 and 10 violations at the three densities against
    24-31 for the single TC-CBF and 60-80 for the ED-CBF.}
    \label{fig:severity}
\end{figure}

The severity of the residual failures further distinguishes the methods (Fig.~\ref{fig:severity}). When the proposed planner violates the safety zone, the penetration remains relatively small. Across all traffic densities, the median penetration is 44 m for a combined safety radius of approximately 1 km, with 14 violations in 300 trials. At the highest density, the median and maximum penetrations are 19 m and 106 m, respectively. In comparison, the single TC-CBF yields median and maximum penetrations of 605 m and 890 m, while the undirected ED-CBF shows an intermediate median penetration of 250 m. These results indicate that topology enumeration reduces not only the frequency of safety-zone violations but also their severity relative to the single-mode baselines.

\begin{table}[tbp]
\centering
\footnotesize
\caption{Per-period solution time versus branching capacity for a fixed
constraint budget of $N_o=6$ (Intel Core Ultra~7 270K Plus; mean $\pm$
standard deviation over 50 runs).}
\label{tab:scaling}
\setlength{\tabcolsep}{3.2pt}
\renewcommand{\arraystretch}{1.3}
\begin{tabular}{cccccc}
\toprule
$M$ & $K=2^{M}$ & \makecell{\textbf{Sequential}\\ (ms)} &
\makecell{\textbf{Parallel}\\ (ms)} & \makecell{\textbf{Speed-}\\ \textbf{up}} &
\textbf{Threads}\\
\midrule
1 & 2 & $0.85\pm0.13$ & $0.49\pm0.10$ & 1.7$\times$ & 2\\
2 & 4 & $1.69\pm0.16$ & $0.54\pm0.13$ & 3.1$\times$ & 4\\
3 & 8 & $3.45\pm0.29$ & $0.59\pm0.11$ & 5.8$\times$ & 8\\
4 & 16 & $8.23\pm0.56$ & $1.35\pm0.19$ & 6.1$\times$ & 12\\
5 & 32 & $16.00\pm0.65$ & $2.26\pm0.25$ & 7.1$\times$ & 12\\
\bottomrule
\end{tabular}
\end{table}

\subsection{Computational Cost}\label{sec:results_scaling}
Table~\ref{tab:scaling} evaluates the computational cost of mode enumeration
for a representative six-ship encounter on a 24-core CPU, using 50
repetitions after warm-up. The total number of considered ships is fixed at
$N_o=6$ for all configurations, with $M$ branched ships and $6-M$ guard
ships. Thus, the OCP dimensions remain unchanged, isolating the computational
effect of increasing the number of modes.

The sequential solution time increases approximately linearly with
$K=2^M$, reaching 16 ms at $K=32$. In contrast, OpenMP-based parallel
execution reduces the batch time to below 2.3 ms for all tested
configurations, with a maximum speed-up of $7.1\times$. The parallel cost
increases primarily with the number of dispatch waves, approximately
following $\lceil K/P\rceil$ for $P$ available threads, rather than directly
with $K$.

In closed-loop simulations at the highest traffic density \textbf{D3}, the
proposed method requires an average batch solution time of 1.7 ms per
replanning period, with a worst-case time below 4 ms. These values are well
within the 1 s replanning period. The remaining serial overhead, including
parameter updates and mode selection, is below 0.15 ms. Overall, the
parallel implementation enables multiple avoidance topologies to be
evaluated with only a small increase in wall-clock computation time.
The substantial margin relative to the replanning period also suggests that the framework may remain practical on processors with more limited computational capability, although its performance on such hardware is not evaluated here.

\begin{figure*}[tbp]
    \centering
    \includegraphics[width=\linewidth]{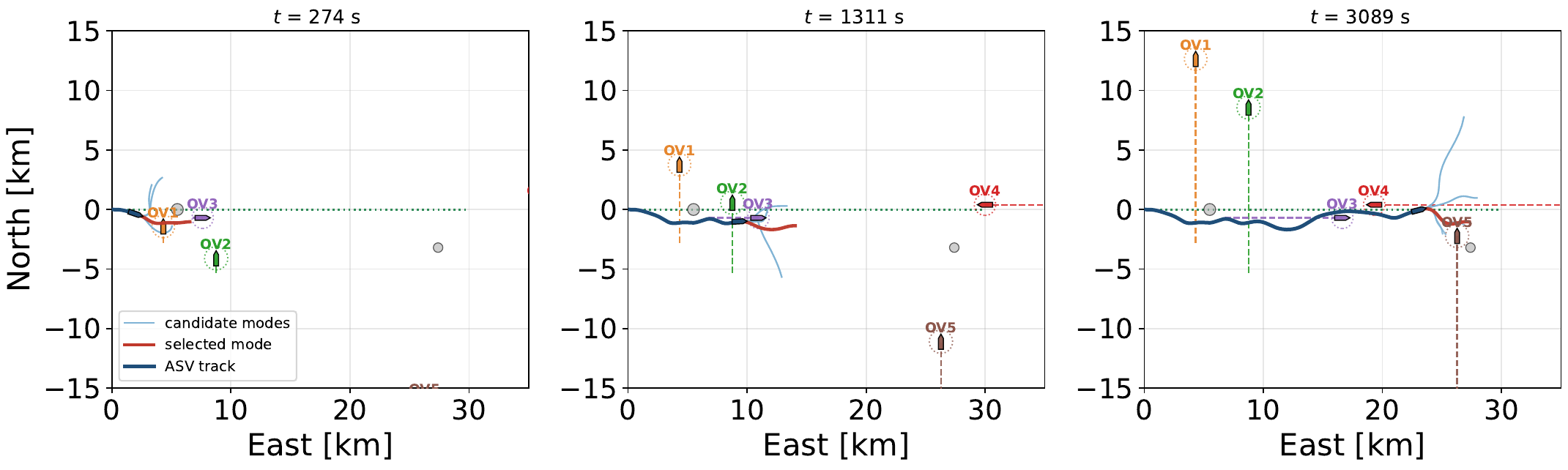}
    \caption{COLREGs-aware mode selection on a straight route with five
    target vessels (OV1-OV5) and two stationary obstacles (gray). Three representative instants are shown on identical axes. Light blue denotes the $K=8$ candidate modes solved in parallel, red the selected mode, and dark blue the realized trajectory. Candidate trajectories with large deviations correspond to unfavorable passing-side combinations and are rejected due to their substantially higher costs.}
    \label{fig:colregs}
\end{figure*}

\subsection{Extension: Navigation-Rule Preferences in the Mode Selection} \label{sec:results_colregs} Because every mode is solved before a decision is made, the mode-selection stage provides a natural point for introducing navigation-rule preferences \cite{COLREG}. For mode $k$, let \begin{equation}\label{eq:rule_violation_set} \mathcal{V}_k=\left\{j\;\middle|\;\sigma_j^{(k)}\neq\sigma_j^{\mathrm{pref}}\right\} \end{equation} denote the set of obstacles for which the side assignment of mode $k$ differs from the rule-preferred side. The rule-adjusted cost is defined as \begin{equation}\label{eq:colregs_cost} \widetilde{J}_k=J_k^{\ast}+\lambda\,\bar{J}\sum_{j\in\mathcal{V}_k}w_j u_j, \end{equation} where $\sigma_j^{\mathrm{pref}}$ is the passing side preferred by the rule applicable to obstacle $j$, $w_j$ specifies the strength of the preference, $u_j\in[0,1]$ represents encounter urgency based on DCPA and TCPA, and $\bar{J}$ is the median feasible cost used to scale the penalty. The encounter type follows a standard DCPA/TCPA classification \cite{pvo_cho2020efficient}. Let $\hat{k}=\argmin_{k\in\mathcal{K}_f}\widetilde{J}_k$ denote the feasible mode with the lowest rule-adjusted cost. The selected mode is then determined by \begin{equation}\label{eq:colregs_prior} k^\ast=\begin{cases} \hat{k}, & \text{if } k_{\mathrm{prev}}\notin\mathcal{K}_f \text{ or } \widetilde{J}_{\hat{k}}+\eta\bar{J}<\widetilde{J}_{k_{\mathrm{prev}}},\\ k_{\mathrm{prev}}, & \text{otherwise}. \end{cases} \end{equation} 
Thus, a mode switch occurs only when the reduction in the rule-adjusted cost
exceeds the threshold $\eta\bar{J}$. Importantly, the underlying OCPs remain
unchanged: the same $K$ candidate modes are solved regardless of the
navigation-rule preferences, and the rule information affects only the
subsequent mode selection. Because the preferences are imposed as soft
penalties rather than hard constraints, alternative modes remain available
when the rule-preferred topology is infeasible or excessively costly.

The extension is demonstrated on a straight 30\,km route with five
non-reactive target vessels and two stationary obstacles. The encounters
include three crossing vessels from starboard, one overtaking vessel, and
one head-on vessel, corresponding to Rules 15, 13, and 14, respectively.
The planner uses
$M=3$ branched obstacles, yielding $K=8$ candidate modes, together with
three guard constraints.
All five encounters are resolved consistently with the applicable navigation
rules, with no safety-zone violation. The ego ship passes astern of the
crossing vessels, maintains clearance during overtaking, and performs a
port-to-port passage in the head-on encounter. 

The soft rule preference can alter the selected topology even when another mode has a lower nominal trajectory cost. Conversely, when compliance with the preferred side incurs an excessively large cost, the penalty can be outweighed and another already-computed feasible mode can be selected. This behavior provides a natural fallback mechanism without modifying or re-solving the underlying OCPs.

Figure~\ref{fig:colregs} illustrates representative encounters together with the candidate trajectories. Some rejected modes exhibit large deviations because their prescribed passing-side combinations are poorly matched to the current geometry, resulting in saturated steering and large barrier slack. Their correspondingly high costs cause them to be rejected by the mode selector. Candidates associated with inactive side constraints may coincide, so fewer than eight distinct trajectories can be visually apparent.

\section{Conclusion}\label{sec:conclusion}
This paper presented a guide path-free multimodal trajectory planning framework for surface vehicles, in which avoidance topologies are encoded directly through the side parameters of turning circle-based CBFs. Enumerating the port/starboard assignments of the nearest $M$ ships yields $2^M$ structurally identical OCPs, while additional ships are handled by fixed-side guard constraints. This removes the need for a separate high-level planner or guide-path initialization. The TC-CBF accounts for finite turning capability through the minimum turning radius, while the MPC incorporates first-order Nomoto dynamics. The common solver structure enables efficient parallel evaluation, solving up to 32 modes in less than 2.5 ms per replanning period. Monte Carlo simulations showed that the $K=16$ configuration achieved success rates of 99\%, 97\%, and 86\%, outperforming the single-mode baselines while substantially reducing safety-zone penetration. These results demonstrate improved robustness and safety in multi-ship encounters with real-time computational feasibility. Future work will consider interaction-aware traffic prediction and prediction uncertainty.

\section*{Declaration of generative AI and AI-assisted technologies in
the manuscript preparation process}
During the preparation of this work the author used ChatGPT in order to
improve the readability of the manuscript and to assist with the
implementation of the simulation environment. After using this tool, the
author reviewed and edited the content as needed and takes full
responsibility for the content of the published article.

\printcredits

\bibliographystyle{cas-model2-names}
\bibliography{ref}

\balance

\bio{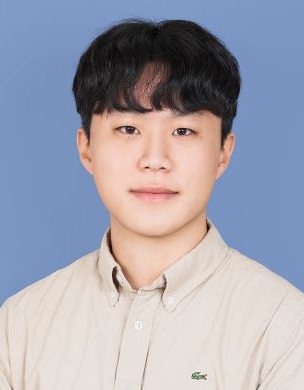}
\textbf{Changyu Lee}
received the B.S. degree in mathematics from Hanyang University, Seoul,
South Korea, in 2018, and the M.S. and Ph.D. degrees in mechanical
engineering from the Korea Advanced Institute of Science and Technology
(KAIST), Daejeon, South Korea, in 2020 and 2025, respectively. He is an
Assistant Professor with the Department of Mechanical and Automotive
Engineering, Kongju National University, Cheonan, South Korea. His research
interests include trajectory planning and robust control.
\endbio

\end{document}